\PassOptionsToPackage{unicode}{hyperref}
\PassOptionsToPackage{hyphens}{url}
\documentclass[
]{article}
\usepackage{xcolor}
\usepackage{amsmath,amssymb}
\usepackage{iftex}
\ifPDFTeX
  \usepackage[T1]{fontenc}
  \usepackage[utf8]{inputenc}
  \usepackage{textcomp} 
\else 
  \usepackage{unicode-math} 
  \defaultfontfeatures{Scale=MatchLowercase}
  \defaultfontfeatures[\rmfamily]{Ligatures=TeX,Scale=1}
\fi
\usepackage{lmodern}
\ifPDFTeX\else
\fi
\IfFileExists{upquote.sty}{\usepackage{upquote}}{}
\IfFileExists{microtype.sty}{
  \usepackage[]{microtype}
  \UseMicrotypeSet[protrusion]{basicmath} 
}{}
\makeatletter
\@ifundefined{KOMAClassName}{
  \IfFileExists{parskip.sty}{%
    \usepackage{parskip}
  }{
    \setlength{\parindent}{0pt}
    \setlength{\parskip}{6pt plus 2pt minus 1pt}}
}{
  \KOMAoptions{parskip=half}}
\makeatother
\usepackage{longtable,booktabs,array}
\usepackage{calc} 
\usepackage{etoolbox}
\makeatletter
\patchcmd\longtable{\par}{\if@noskipsec\mbox{}\fi\par}{}{}
\makeatother
\IfFileExists{footnotehyper.sty}{\usepackage{footnotehyper}}{\usepackage{footnote}}
\makesavenoteenv{longtable}
\usepackage{graphicx}
\makeatletter
\newsavebox\pandoc@box
\newcommand*\pandocbounded[1]{
  \sbox\pandoc@box{#1}%
  \Gscale@div\@tempa{\textheight}{\dimexpr\ht\pandoc@box+\dp\pandoc@box\relax}%
  \Gscale@div\@tempb{\linewidth}{\wd\pandoc@box}%
  \ifdim\@tempb\p@<\@tempa\p@\let\@tempa\@tempb\fi
  \ifdim\@tempa\p@<\p@\scalebox{\@tempa}{\usebox\pandoc@box}%
  \else\usebox{\pandoc@box}%
  \fi%
}
\def\fps@figure{htbp}
\makeatother
\providecommand{\tightlist}{%
  \setlength{\itemsep}{0pt}\setlength{\parskip}{0pt}}
\usepackage[margin=1in]{geometry}
\usepackage{float}
\usepackage{bookmark}
\IfFileExists{xurl.sty}{\usepackage{xurl}}{} 
\hypersetup{
  pdftitle={The Geno-Synthetic Algorithm: Type-Factored Coevolutionary Optimization for Heterogeneous Genotypes and Assembled Phenotypes},
  pdfkeywords={evolutionary computation, differential
evolution, cooperative coevolution, mixed-variable
optimization, genotype-phenotype mapping, vector-space
optimization, prompt optimization, embedding optimization, financial
machine learning},
  hidelinks,
  pdfcreator={LaTeX via pandoc}}

\title{The Geno-Synthetic Algorithm: Type-Factored Coevolutionary
Optimization for Heterogeneous Genotypes and Assembled Phenotypes}
\author{Alex Bogdan\\
Evolutionairy AI\\
Toronto, Canada}
\date{May 2026}

\begin{document}
\maketitle
\begin{abstract}
Many real-world optimization problems cannot be naturally represented as
homogeneous real-valued vectors. Instead, they are composite design
objects with parameters such as integers, real numbers, Boolean
switches, categorical selections, complex-valued descriptors, and
increasingly, embedding-space representations. Standard evolutionary
algorithms often flatten these heterogeneous components into a single
chromosome and apply generic variation operators, relying on rounding,
repair, encoding techniques, or post-hoc constraint correction. The
Geno-Synthetic Algorithm (GSA) is introduced as a type-factored
coevolutionary optimization framework. In GSA, genes are partitioned by
representational type or semantic family, evolved in parallel using
type-native evolutionary operators, and synthetically assembled into
complete phenotypes for joint fitness evaluation.

GSA was initially developed for a machine-learning-based investment
system, WALLACE, in which candidate portfolio-selection models comprised
heterogeneous gene families, including integer lookback periods,
continuous thresholds, volatility and correlation parameters, Boolean
filters, ranking weights, and structurally complex descriptors. The
central insight is that the geometry of each gene should determine the
evolutionary operator applied to it. Integer-, real-valued-, Boolean-,
and complex-valued genes, as well as embedding vectors, possess distinct
topologies, mutation semantics, and admissible transformations.
Consequently, GSA avoids treating a heterogeneous design object as a
single undifferentiated vector. Instead, each gene family evolves within
its native representational space, and the resulting sub-genomes are
assembled into richer, executable phenotypes.

This paper formalizes GSA as a typed product-space search procedure with
an explicit phenotype assembly operator. The method is positioned
relative to differential evolution, mixed-variable evolutionary
algorithms, cooperative coevolution, and genotype-phenotype mapping. An
open-source reference implementation (\texttt{gsa-experiments},
available at
\url{https://github.com/Wallace-AI/Geno_Synthetic_Algorithm}) is
released, and a focused empirical evaluation is reported across seven
benchmark problems, six synthetic and the external COCO BBOB-MixInt
suite, comparing eight GSA variants (including synchronous and
asynchronous schedules) against five established baselines, with
evaluation budgets ranging from 5,000 to 100,000. The primary empirical
finding is architectural: GSA is the only method that operates when gene
families include complex-valued descriptors or embedding vectors, while
remaining competitive on Boolean-only and single-family problems. On
smooth synthetic multi-family problems, well-tuned flattened
differential evolution remains the strongest baseline at all tested
budgets. On the external COCO BBOB-MixInt suite, performance varies with
budget: at 5,000 evaluations, FLATTENED\_DE is superior, but at 100,000
evaluations, GSA\_DIRECT is statistically indistinguishable from
FLATTENED\_DE (\(\hat{A}_{12}\) = 0.499, p = 0.61), and FLATTENED\_EA
drops from second to fifth rank, demonstrating the predicted asymptotic
crossover. These results clarify GSA\textquotesingle s value:
architectural reach across families that flattened encodings cannot
represent, parity with the strongest flattened baseline at COCO-scale
budgets, superiority over EA-style flattened baselines at larger
budgets, and graceful degradation rather than catastrophic failure on
benchmarks where differential evolution is near-optimal at small
budgets. Four ablation findings further refine these conclusions:
type-native operators are essential within the architecture;
ensemble-context credit is consistently dominated by elite credit;
active phenotype assembly is statistically favored over passive
concatenation on benchmarks requiring gating (p \(\approx\)
3·\(10^{-7}\) across D \(\in\) \{20, 40, 80\}); and asynchronous
schedules, while supported by the architecture, are not optimal at fixed
evaluation budgets and should be reserved for scenarios with
heterogeneous per-family evaluation costs. Finally, the architecture is
argued to extend naturally beyond finance to modern
representation-learning problems, including optimization over soft
prompts, concept vectors, symbolic-control parameters, and hybrid latent
spaces in large language model systems.
\end{abstract}


\section{1. Introduction}\label{introduction}

Evolutionary optimization algorithms are often described in simplified
terms, wherein a population of candidate solutions is varied, evaluated,
selected, and iteratively improved. This abstraction enables the
application of a general algorithmic template across diverse domains
such as engineering, machine learning, finance, design, scheduling, and
control. However, this abstraction may obscure a practical challenge
encountered in real systems: the objects being optimized are rarely
homogeneous vectors.

Canonical evolutionary algorithms typically assume that candidate
solutions can be represented as chromosomes composed of comparable
components. For instance, in real-valued differential evolution, a
candidate is represented as a vector in a continuous search space. In
classical genetic algorithms, candidates may be encoded as bit strings,
symbolic chromosomes, or permutations. These representations are
effective when the problem domain aligns with the chosen encoding.
However, many high-value optimization problems are not naturally
Boolean, purely real-valued, purely integer-valued, or purely
categorical; instead, they are heterogeneous design objects.

A machine-learning-based investment system exemplifies this
heterogeneity. A candidate portfolio-selection model may include integer
lookback periods, real-valued weights, volatility thresholds,
correlation coefficients, Boolean switches to enable or disable filters,
categorical choices among indicators, and complex-valued descriptors
that capture phase-like, spectral, or fractal structure. Similarly, a
language-model prompt optimization system may comprise discrete textual
instructions, continuous soft-prompt vectors, Boolean routing controls,
categorical tool choices, and real-valued decoding parameters. A
robotics controller may integrate real-valued gains, discrete modes,
symbolic rules, and learned latent representations. In each of these
cases, the final executable system is assembled from components
belonging to distinct representational spaces.

A common engineering approach involves flattening the entire candidate
into a single chromosome and applying a generic evolutionary procedure.
Continuous values are rounded to integers, Boolean values are
thresholded, categorical values are numerically encoded, and invalid
solutions are repaired after mutation. Although convenient, this
approach lacks representational fidelity. Boolean switches do not mutate
in the same manner as floating-point coefficients. Integer lookback
periods possess ordered but discrete geometry. Complex-valued
descriptors may encode both magnitude and phase. Embedding vectors may
reside on or near learned semantic manifolds where Euclidean
perturbation is not always meaningful. Treating these diverse components
as interchangeable coordinates can result in invalid mutations,
inefficient search, brittle convergence, and loss of valuable structural
information.

The Geno-Synthetic Algorithm (GSA) is introduced as a type-factored
evolutionary optimization framework specifically designed for
heterogeneous design objects. The central concept is to partition the
genotype into homogeneous or semantically coherent gene families, evolve
each family within its native representational space, and then
synthetically assemble the evolved sub-genomes into complete candidate
phenotypes for fitness evaluation, rather than evolving a single
flattened chromosome.

The term Geno-Synthetic is used intentionally. The algorithm is
described as Geno- because it operates on genotype components, which are
separate families of encoded parameters. It is synthetic because the
candidate solution is not merely the concatenation of these components.
The meaningful entity is the assembled phenotype, produced by a
synthesis function that maps multiple typed sub-genomes into an
executable model, strategy, controller, prompt, or decision system.

The practical motivation for GSA originated from the development of
WALLACE, a machine-learning-based investment system named in honor of
Alfred Russel Wallace. In this context, candidate models consisted of
multiple distinct parameter families. Evolving all genes within a single
homogeneous chromosome introduced unnecessary complexity and
instability. Partitioning the representation by data type and semantic
function resulted in a cleaner architecture, enabled independent
evolutionary processes to operate in parallel, improved convergence
robustness, and facilitated the construction of richer phenotypes.

The broader claim of this paper is not that GSA replaces differential
evolution, genetic algorithms, cooperative coevolution, or
mixed-variable evolutionary algorithms. Rather, GSA identifies and
formalizes a key design principle: in heterogeneous optimization
problems, the representational geometry of each gene family should
determine the evolutionary process applied. A typed genotype should not
be forced into a single artificial space when its components possess
different mathematical structures, mutation semantics, and constraints.

The contributions of this paper are summarized as follows:

1. Introduction of GSA as a type-factored coevolutionary framework for
mixed-representation optimization.

2. Formalization of GSA using typed product spaces, type-native
evolutionary operators, and an explicit phenotype assembly function.

3. Distinction of GSA from conventional differential evolution,
mixed-variable evolutionary algorithms, and cooperative coevolution.

4. Description of the original financial machine-learning motivation
that led to the algorithm.

5. Release of an open-source reference implementation
(\href{https://github.com/Wallace-AI/Geno_Synthetic_Algorithm}{\texttt{gsa-experiments}},
MIT-licensed) and a reproducible empirical battery comparing six GSA
variants against five baselines on five typed synthetic benchmarks plus
OneMax.

6. Outline of how the same framework may be applied to embedding space
and prompt optimization for large language model systems.

\begin{figure}
\centering
\includegraphics[width=1\linewidth,height=\textheight,keepaspectratio,alt={Canonical genetic algorithm (left) versus the Geno-Synthetic Algorithm (right). The canonical approach flattens heterogeneous parameters into a single chromosome and applies undifferentiated variation, sacrificing representational fidelity for gene families that are not naturally real-valued. GSA partitions the genotype by representational type, evolves each family in parallel with type-native operators, and synthesises a candidate phenotype through an explicit assembly operator.}]{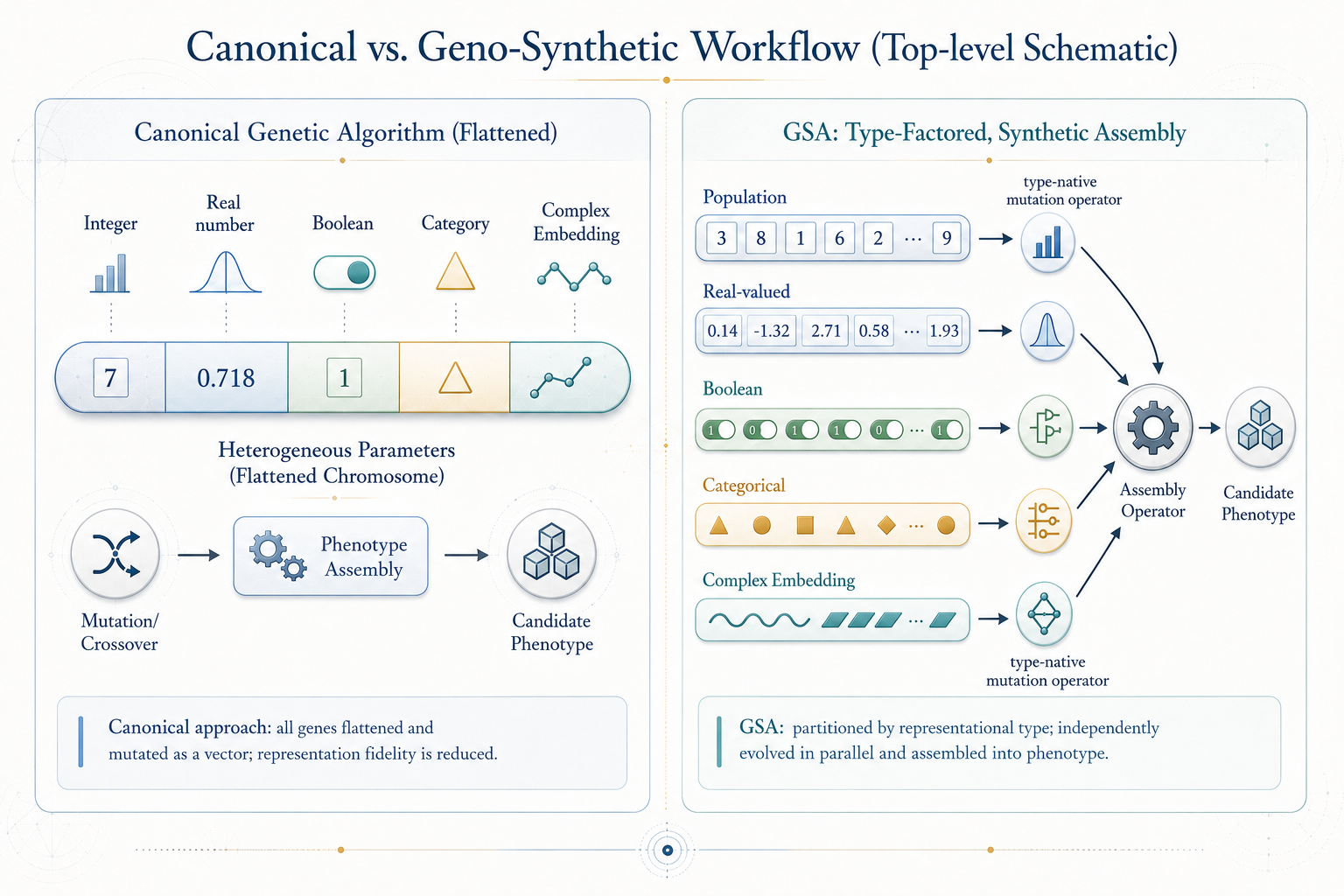}
\caption{Canonical genetic algorithm (left) versus the Geno-Synthetic
Algorithm (right). The canonical approach flattens heterogeneous
parameters into a single chromosome and applies undifferentiated
variation, sacrificing representational fidelity for gene families that
are not naturally real-valued. GSA partitions the genotype by
representational type, evolves each family in parallel with type-native
operators, and synthesises a candidate phenotype through an explicit
assembly operator.}
\end{figure}

The remainder of the paper is organized as follows:

Section 2 reviews relevant background.

Section 3 describes the motivating problem of heterogeneous genotypes in
real-world model construction.

Section 4 formalizes the GSA framework.

Section 5 presents the algorithmic procedure.

Section 6 discusses expected advantages.

Section 7 addresses limitations.

Section 8 reports the empirical evaluation.

Section 9 discusses extensions to financial machine learning and large
language model systems.

Section 10 concludes.

Section 11 details code and data availability.

\section{2. Background and Related
Work}\label{background-and-related-work}

\subsection{2.1 Differential Evolution}\label{differential-evolution}

Differential evolution (DE) is a population-based global optimization
algorithm initially developed for continuous search spaces. In this
approach \hyperref[ref-storn1997]{(Storn and Price, 1997)}, a population
of real-valued vectors evolves through difference-based mutation,
crossover, and selection. Alongside CMA-ES
\hyperref[ref-hansen2001]{(Hansen and Ostermeier, 2001)}, DE remains one
of the strongest baselines for continuous-vector black-box optimization.
In the canonical DE algorithm \hyperref[ref-storn1997]{(Storn and Price,
1997)}, a mutant vector is generated by adding a scaled difference
between two population members to a third vector. This mutant is then
recombined with a target vector, and the superior candidate is retained
following fitness evaluation.

The primary advantages of DE are its simplicity, robustness, and the few
control parameters required. Its inherent use of real-valued vectors is
advantageous for naturally continuous problems. However, for problems
involving integers, Boolean variables, categorical data, structured
objects, or embeddings with non-Euclidean constraints, this
representation is less suitable. Extensions such as mixed-integer and
mixed-variable DE have been developed to address these limitations
\hyperref[ref-price2005]{(Price et al., 2005)}, although many of these
methods still rely on encoding heterogeneous components into a single
unified chromosome.

The Geno-Synthetic Algorithm (GSA) is compatible with
differential-evolution-style operators when appropriate. For gene
families represented by real-valued data, the DE operator may be the
most suitable evolutionary mechanism. Unlike traditional DE, GSA does
not require all gene families to be treated as residing in the same
real-valued space.

\subsection{2.2 Mixed-Variable Evolutionary
Optimization}\label{mixed-variable-evolutionary-optimization}

Mixed-variable optimization addresses problems involving variables from
distinct domains, including continuous, integer, ordinal, nominal, and
binary spaces. Evolutionary approaches to mixed-variable optimization
frequently employ hybrid encodings, specialized mutation operators,
repair mechanisms, or variable-specific handling rules. These methods,
surveyed in standard treatments of metaheuristics
\hyperref[ref-talbi2009]{(Talbi, 2009;} \hyperref[ref-yang2021]{Yang,
2021)}, are significant and directly applicable to GSA.

In contrast, GSA adopts a distinct organizing principle. Instead of
considering mixed variables as exceptions within a single chromosome,
GSA regards heterogeneity as the fundamental structure of the problem.
Each gene family evolves as an independent sub-genome, each with its own
population, operators, constraints, and potentially its own evolutionary
tempo. The final solution is constructed through phenotype synthesis.

\subsection{2.3 Cooperative Coevolution}\label{cooperative-coevolution}

Cooperative coevolution \hyperref[ref-potter1994]{(Potter and De Jong,
1994)} decomposes high-dimensional problems into subcomponents, each of
which is optimized by a distinct subpopulation. Candidates from one
subpopulation are evaluated alongside representatives from others. This
divide-and-conquer strategy is particularly effective for large-scale
optimization, especially when subcomponent grouping is learned from
variable-interaction structure \hyperref[ref-omidvar2014]{(Omidvar et
al., 2014)}.

GSA shares significant similarities with cooperative coevolution, as
both approaches utilize subcomponent evolution and joint evaluation.
However, the primary distinction lies in their decomposition principles.
Cooperative coevolution typically decomposes problems based on variable
grouping, separability, interaction structure, random partitioning, or
learned decomposition. In contrast, GSA initially decomposes by
representational type and, when appropriate, by semantic gene family.
This approach assumes that data type and semantic role frequently encode
valuable prior knowledge regarding mutation validity, distance,
recombination, and constraint structure.

This distinction is significant. For example, a group of integer
lookback-period genes may interact closely with real-valued thresholds,
yet it can still be advantageous to evolve the integer gene family using
integer-specific operators and the threshold family with continuous
operators. GSA accommodates cross-family interactions through phenotype
assembly and a joint fitness feedback loop.

\subsection{2.4 Genotype-Phenotype
Mapping}\label{genotype-phenotype-mapping}

In classical evolutionary computation
\hyperref[ref-holland1975]{(Holland, 1975;}
\hyperref[ref-mitchell1998]{Mitchell, 1998;}
\hyperref[ref-back1996]{Bäck, 1996)}, the genotype differs from the
phenotype. The phenotype results from developmental processes,
regulatory interactions, environmental conditions, and expression
dynamics. Evolutionary computation frequently simplifies this
relationship by mapping the encoded chromosome directly to the candidate
solution. However, in many engineering applications, a more nuanced
genotype-phenotype distinction provides significant advantages.

GSA conceptualizes the phenotype as an assembled system rather than a
raw chromosome. The genotype comprises type-factored sub-genomes. An
assembly operator produces the phenotype by combining, validating,
transforming, and activating these sub-genomes into an executable
candidate. Consequently, the assembly operator becomes a central focus
of study rather than a minor implementation detail.

\subsection{2.5 Evolutionary Prompt and Embedding
Optimization}\label{evolutionary-prompt-and-embedding-optimization}

Recent advancements in large language models have revitalized research
in prompt optimization \hyperref[ref-lester2021]{(Lester et al., 2021;}
\hyperref[ref-guo2023]{Guo et al., 2023)}. Prompts may be optimized as
discrete natural-language objects, continuous soft-prompt embeddings, or
hybrid structures that incorporate both symbolic and latent components.
Evolutionary algorithms are particularly suitable for this context
because they enable optimization of black-box systems without requiring
gradient information, complementing surrogate-gradient techniques such
as the straight-through estimator \hyperref[ref-bengio2013]{(Bengio et
al., 2013)}.

Prompt optimization is inherently heterogeneous. A comprehensive large
language model (LLM) control object may include discrete instructions,
examples, role definitions, tool-selection policies, retrieval settings,
decoding parameters, safety constraints, routing rules, and embedding
vectors. Consequently, the prompt-optimization domain is well-suited for
GSA-style type-factored evolutionary approaches.

\subsection{2.6 Novelty and Positioning of the Geno-Synthetic
Algorithm}\label{novelty-and-positioning-of-the-geno-synthetic-algorithm}

The Geno-Synthetic Algorithm (GSA) represents an advancement in the
representation and architecture of evolutionary search, rather than
asserting that heterogeneous variables have not previously been
addressed. Mixed-variable evolutionary algorithms, mixed-integer
differential evolution, genetic algorithms with hybrid encodings, and
cooperative coevolutionary systems all address related challenges. GSA
is situated within this broader family of methods, yet its organizing
principle remains distinct.

The primary innovation of GSA is type-factored phenotype synthesis. In
this approach, the decomposition of candidate solutions is not primarily
motivated by dimensionality reduction, random grouping, separability
detection, or convenience of encoding. Rather, decomposition is
determined by the genes\textquotesingle{} representational type and
semantic role. Integer, real-valued, Boolean, and categorical genes, as
well as complex-valued descriptors and embedding-space variables, are
each assigned to distinct evolutionary geometries. Each gene family
evolves using operators suited to its geometry, and complete candidate
solutions are generated only after a synthetic assembly step.

This distinction is significant because many optimization procedures
conceal heterogeneity within a unified encoding. For instance, integer
genes may be represented as floating-point coordinates and rounded after
mutation, Boolean genes as scalars and thresholded, categorical
variables by arbitrary numerical codes, and structured descriptors may
be flattened into coordinates that lose their original meaning. While
these techniques are often practical, they shift the burden of
representation from the evolutionary process to encoding and repair. In
contrast, GSA treats heterogeneity as an inherent and natural structure
of the problem, rather than as an inconvenience.

\subsubsection{2.6.1 Difference from Canonical Differential
Evolution}\label{difference-from-canonical-differential-evolution}

Canonical differential evolution is primarily defined over continuous
real-valued vectors. The mutation operator depends on scaled differences
between candidate vectors, which is only meaningful when the search
space exhibits continuous vector geometry. While GSA can employ
differential-evolution-style operators for real-valued sub-genomes, it
does not require that all components of the genotype reside within the
same continuous space.

Within GSA, a complete candidate may include continuous coefficients,
discrete lookback windows, Boolean activation masks, categorical feature
selections, complex-valued structural descriptors, and embedding
vectors. Applying a single differential operator to the entire flattened
candidate would incorrectly impose a uniform geometry. GSA addresses
this by applying differential or continuous operators only to suitable
subcomponents and employing alternative operators for other gene
families.

Therefore, GSA does not reject differential evolution. Instead, it
generalizes evolutionary search architecture, allowing differential
evolution to function as one type-native operator among several
alternatives.

\subsubsection{2.6.2 Difference from Mixed-Variable Differential
Evolution}\label{difference-from-mixed-variable-differential-evolution}

Mixed-variable differential evolution extends differential evolution
(DE) to address problems involving continuous, integer, discrete, or
categorical variables. This constitutes highly relevant prior work.
However, most mixed-variable approaches continue to operate under the
assumption of a single candidate chromosome, introducing specialized
mechanisms to handle different variable types within it.

The Geno-Synthetic Algorithm (GSA) differs from these approaches in four
key aspects.

First, GSA elevates type partitioning to the algorithm\textquotesingle s
primary architectural principle. Typed sub-genomes are not simply fields
within a shared chromosome; instead, they function as independently
evolved subpopulations.

Second, GSA treats each gene family as possessing distinct evolutionary
dynamics. Parameters such as mutation rate, recombination logic,
selection pressure, diversity preservation, and updating frequency may
vary across integer, real-valued, Boolean, categorical, complex-valued,
and embedding-space components.

Third, GSA positions the assembly operator as a central component of the
algorithm. The phenotype is not merely a decoded chromosome; rather, it
is a synthesized executable object generated by combining typed
sub-genomes in accordance with domain-specific structural rules.

Fourth, GSA enables different sub-genomes to evolve either in parallel
or asynchronously. This capability is particularly significant when gene
families exhibit varying evaluation costs, stability properties, or
rates of beneficial change.

\subsubsection{2.6.3 Difference from Cooperative
Coevolution}\label{difference-from-cooperative-coevolution}

GSA shares similarities with cooperative coevolution, as both
methodologies optimize partial components and assess them using joint
fitness. Cooperative coevolution is typically motivated by the
challenges associated with optimizing high-dimensional problems. The
primary concern in this context is determining how variables should be
grouped to ensure that interacting variables are optimized collectively
and separability is effectively utilized.

In contrast, GSA focuses on identifying the native representational
structure of the genes undergoing evolution.

In cooperative coevolution, decomposition frequently relies on
interaction structure, separability assumptions, random grouping,
differential grouping, or learned linkage. Conversely, GSA bases its
initial decomposition on data type and semantic gene family. As a
result, even when integer lookbacks interact significantly with
real-valued thresholds, these components may evolve into distinct typed
subpopulations due to differing mutation semantics. Their interaction is
managed through phenotype assembly and joint fitness feedback, rather
than by combining them into a single homogeneous chromosome.

This characteristic renders GSA particularly suitable for systems in
which the candidate solution is a structured artifact composed of
diverse decision elements, rather than a simple high-dimensional vector.
Relevant examples include investment-selection systems, hybrid
machine-learning pipelines, symbolic-continuous controllers, and
prompt-optimization systems that integrate text, embeddings, Boolean
routing rules, and scalar parameters.

\subsubsection{2.6.4 Difference from Standard Genotype-Phenotype
Mapping}\label{difference-from-standard-genotype-phenotype-mapping}

Many evolutionary algorithms include a decoding step that maps genotype
to phenotype. In the Geno-Synthetic Algorithm (GSA), this mapping is
more explicit and central to the process. The assembly operator
functions as an active synthesis mechanism, combining typed sub-genomes
into a valid candidate system rather than serving as a passive decoder.

For example, in a financial machine-learning system, Boolean genes
determine which feature families are expressed, integer genes specify
lookback horizons, real-valued genes assign feature weights, and
structural genes define higher-order descriptors. The resulting
phenotype is a functioning portfolio-selection model rather than a
simple list of parameter values.

This distinction is significant because the expressive power of GSA
arises from the interaction between typed evolution and phenotype
assembly. The assembly operator enables the algorithm to construct
complex behavioral systems from simpler, type-specific evolutionary
processes.

\subsubsection{2.6.5 Difference from Evolutionary Prompt
Optimization}\label{difference-from-evolutionary-prompt-optimization}

Recent evolutionary prompt-optimization methods
\hyperref[ref-guo2023]{(Guo et al., 2023)} apply evolutionary principles
to enhance natural-language prompts, typically treating prompts as
discrete textual entities. While GSA can incorporate discrete prompt
evolution, its broader significance lies in optimizing hybrid
LLM-control objects. A practical LLM configuration may comprise
natural-language instructions, few-shot examples, tool-use switches,
retrieval parameters, decoding coefficients, safety gates, memory
controls, and soft-prompt or concept-vector embeddings.

GSA provides effective architecture for hybrid optimization by evolving
each component within its native space and evaluating the assembled
LLM-control phenotype based on task performance, cost, reliability, or
safety.

\subsubsection{2.6.6 Summary of
Distinctions}\label{summary-of-distinctions}

\begin{longtable}[]{@{}
  >{\raggedright\arraybackslash}p{(\linewidth - 4\tabcolsep) * \real{0.2692}}
  >{\raggedright\arraybackslash}p{(\linewidth - 4\tabcolsep) * \real{0.3333}}
  >{\raggedright\arraybackslash}p{(\linewidth - 4\tabcolsep) * \real{0.3718}}@{}}
\toprule\noalign{}
\begin{minipage}[b]{\linewidth}\raggedright
Prior approach
\end{minipage} & \begin{minipage}[b]{\linewidth}\raggedright
Typical organizing principle
\end{minipage} & \begin{minipage}[b]{\linewidth}\raggedright
GSA distinction
\end{minipage} \\
\midrule\noalign{}
\endhead
\bottomrule\noalign{}
\endlastfoot
Canonical differential evolution & Continuous vector-space optimization
& Uses DE only where continuous vector geometry is appropriate \\
Mixed-variable DE & Single chromosome with mixed-variable handling &
Uses separate typed subpopulations and type-native evolutionary
dynamics \\
Cooperative coevolution & Divide-and-conquer decomposition of
high-dimensional problems & Decomposes initially by representational
type and semantic gene family \\
Standard genotype-phenotype mapping & Decode the chromosome into a
candidate solution & Treats assembly as an active synthesis operator \\
Evolutionary prompt optimization & Evolve discrete prompts or prompt
populations & Extends naturally to hybrid text, control, and
embedding-space phenotypes \\
\end{longtable}

The defensible novelty claim is therefore not that GSA is the first
evolutionary algorithm to handle mixed variables or decomposed
optimization. Rather, GSA contributes a coherent architecture for
\textbf{typed evolutionary search over heterogeneous genotypes with
explicit phenotype synthesis}.

\section{3. Motivation: Heterogeneous Genotypes in Real
Systems}\label{motivation-heterogeneous-genotypes-in-real-systems}

The motivation for GSA arises from a practical challenge: constructing
and evolving machine-learning-based investment-selection systems. In
these systems, candidate models are seldom characterized by a single
type of parameter. Instead, they are structured decision objects
composed of multiple feature families and control mechanisms.

For instance, a portfolio-selection model may incorporate the following
components:

\begin{itemize}
\item
  lookback periods for momentum, volatility, correlation, and
  mean-reversion measures;
\item
  continuous coefficients controlling the relative influence of feature
  families;
\item
  thresholds for activating or suppressing filters;
\item
  Boolean flags that enable or disable rules;
\item
  categorical choices among feature definitions;
\item
  ranking weights used to combine feature ranks;
\item
  volatility regime controls;
\item
  market-bias parameters;
\item
  correlation, beta, or independence constraints;
\item
  complex-valued or spectral descriptors used to represent structural
  properties of price behavior;
\item
  model-selection or ensemble-selection switches.
\end{itemize}

A flattened chromosome can encode all these values, but this approach
introduces semantic distortion. Mutations that are appropriate for
continuous thresholds may be irrelevant for Boolean filters. Similarly,
crossovers suitable for vectors of real-valued weights may yield
incoherent combinations when indiscriminately applied across indicator
definitions, lookback windows, and regime switches. Although repair
functions can mitigate invalidity, they do not address the fundamental
issue: the representation fails to respect the inherent structure of the
search space.

GSA begins from a different premise. The genotype is partitioned into
type-homogeneous or semantically coherent gene families. For example:

\[G^{\left( \mathbb{Z} \right)} = \left\{ \text{lookback periods, holding periods, rank depths, window size}\text{s} \right\}\]

\[G^{\left( \mathbb{R} \right)} = \left\{ \text{weights, thresholds, scaling coefficients, volatility parameters} \right\}\]

\[G^{\left( \mathbb{B} \right)} = \left\{ \text{feature switches, filter activations, regime flag}\text{s} \right\}\]

\[G^{\left( \mathbb{C} \right)} = \left\{ \text{complex-valued descriptors, spectral coefficients, phase-like variable}\text{s} \right\}\]

\[G^{\left( \mathcal{E} \right)} = \left\{ \text{embedding vectors, soft prompts, latent concept coordinate}\text{s} \right\}\]

The final candidate is assembled from these subgenomes:

\[\theta = \mathcal{A}\left( G^{\left( \mathbb{Z} \right)},G^{\left( \mathbb{R} \right)},G^{\left( \mathbb{B} \right)},G^{\left( \mathbb{C} \right)},G^{\left( \mathcal{E} \right)} \right)\]

Where \(\mathcal{A}\) is the phenotype assembly function.

From this perspective, the algorithm evolves a population of partial
design grammars, each of which manifests as a functional model. The
resulting phenotype may take the form of a portfolio-selection engine,
trading system, control policy, prompt configuration, neural module, or
other executable entity.

A key insight motivating the development of GSA was that type factoring
simplifies implementation and enhances convergence behavior. Integer
lookbacks can evolve independently from floating-point coefficients.
Boolean filter structures can be explored without affecting real-valued
search dynamics. Continuous parameters are optimized using operators
suited for smooth spaces. Complex-valued descriptors retain their
algebraic structure. These independently evolved sub-genomes are
subsequently assembled into candidate phenotypes and evaluated using the
true objective function.

This architecture also facilitates parallelization. Because typed
subpopulations can evolve semi-independently, each subpopulation may be
assigned to a distinct process, thread, GPU kernel, node, or
asynchronous worker. While the fitness signal is generated by fully
assembled candidates, the variation processes can be distributed across
computational resources.

\section{4. Formal Definition of the Geno-Synthetic
Algorithm}\label{formal-definition-of-the-geno-synthetic-algorithm}

\subsection{4.1 Typed Product Search
Space}\label{typed-product-search-space}

Let an optimization problem be defined over a heterogeneous search
space:

\[\mathcal{G} = \mathcal{G}_{1} \times \mathcal{G}_{2} \times \cdots \times \mathcal{G}_{K}\]

where each \(\mathcal{G}_{k}\) is a typed subspace. Examples include:

\[\mathcal{G}_{k} \in \mathbb{Z}^{d_{k}},\mathbb{R}^{d_{k}},\mathbb{B}^{d_{k}},\mathcal{C}^{d_{k}},\mathbb{C}^{d_{k}},\mathcal{E}^{d_{k}},\Pi^{d_{k}}\]

where \(\mathbb{Z}^{d_{k}}\) denotes integer-valued genes,
\(\mathbb{R}^{d_{k}}\) real-valued genes, \(\mathbb{B}^{d_{k}}\) Boolean
genes, \(\mathcal{C}^{d_{k}}\) categorical genes, \(\mathbb{C}^{d_{k}}\)
complex-valued genes, \(\mathcal{E}^{d_{k}}\) embedding-space genes, and
\(\Pi^{d_{k}}\)permutation-valued genes.

A geno-synthetic candidate is a tuple:

\[G_{i} = \left( G_{i}^{(1)},G_{i}^{(2)},\ldots,G_{i}^{(K)} \right)\]

where \(G_{i}^{(k)} \in \mathcal{G}_{k}\).

Unlike a conventional flattened chromosome, \(G_{i}\) is not assumed to
be acted upon by one universal operator. Instead, each subgenome has its
own variation operator, constraint set, distance metric, and selection
policy.

\subsection{4.2 Phenotype Assembly}\label{phenotype-assembly}

The executable candidate is produced by an assembly operator:

\[\theta_{i} = \mathcal{A}\left( G_{i} \right)\]

or, explicitly:

\[\theta_{i} = \mathcal{A}\left( G_{i}^{(1)},G_{i}^{(2)},\ldots,G_{i}^{(K)} \right)\]

The assembly operator may perform several functions:

\begin{enumerate}
\def\labelenumi{\arabic{enumi}.}
\tightlist
\item
  combine typed subgenomes into a complete candidate;
\item
  enforce cross-type constraints;
\item
  activate or deactivate genes according to Boolean switches;
\item
  map latent or complex descriptors into executable parameters;
\item
  resolve incompatibilities among subgenomes;
\item
  instantiate a model, strategy, prompt, controller, or simulation
  object;
\item
  produce the object evaluated by the fitness function.
\end{enumerate}

The fitness of candidate \(i\) is:

\[F_{i} = \mathcal{L}\left( \theta_{i};D \right)\]

where \(\mathcal{L}\) is the objective or loss function and \(D\) is the
evaluation environment, dataset, simulator, market history, validation
task, or benchmark suite.

\subsection{4.3 Type-Native Evolutionary
Operators}\label{type-native-evolutionary-operators}

Each subspace \(\mathcal{G}_{k}\) is associated with a type-native
evolutionary operator:

\[\mathcal{E}_{k}:\mathcal{P}_{k} \times \mathcal{F} \times \Omega_{k} \rightarrow \mathcal{P}_{k}'\]

where \(\mathcal{P}_{k}\) is the population of sub-genomes of type
\(k\), \(\mathcal{F}\) is the fitness feedback derived from assembled
phenotypes, and \(\Omega_{k}\) contains operator-specific parameters.

For example:

\begin{itemize}
\tightlist
\item
  real-valued genes may use differential mutation, Gaussian mutation,
  covariance adaptation, or evolution strategies;
\item
  integer genes may use bounded integer mutation, discrete differential
  steps, or ordinal-preserving crossover;
\item
  Boolean genes may use bit-flip mutation, mask recombination, or
  sparsity-aware activation rules;
\item
  categorical genes may use frequency-informed sampling, replacement
  mutation, or compatibility-aware recombination;
\item
  complex-valued genes may mutate magnitude and phase separately;
\item
  embedding genes may use manifold-aware perturbation, cosine-preserving
  mutation, local semantic drift, or learned basis-vector recombination.
\end{itemize}

The key principle is that each operator should respect the geometry and
semantics of the gene family it modifies.

\subsection{4.4 Fitness Feedback Across
Subpopulations}\label{fitness-feedback-across-subpopulations}

Although subgenomes evolve in separate typed populations, fitness is
evaluated at the assembled phenotype level. This creates a
credit-assignment problem: how should the algorithm assign the fitness
of a complete candidate to its component subgenomes?

GSA can support several feedback schemes.

\subsubsection{Direct Bundle Feedback}\label{direct-bundle-feedback}

Each assembled candidate is formed from a specific tuple of subgenomes.
The resulting fitness is assigned to all participating subgenomes:

\[F\left( G_{i}^{(k)} \right) \leftarrow F\left( \mathcal{A}\left( G_{i}^{(1)},\ldots,G_{i}^{(K)} \right) \right)\]

This is the simplest approach.

\subsubsection{Elite Context Feedback}\label{elite-context-feedback}

A candidate subgenome is evaluated in combination with elite
representatives from other subpopulations:

\[F\left( G_{j}^{(k)} \right) = \mathcal{L}\left( \mathcal{A}\left( E^{(1)},\ldots,E^{(k - 1)},G_{j}^{(k)},E^{(k + 1)},\ldots,E^{(K)} \right);D \right)\]

where \(E^{(m)}\) is an elite or representative subgenome from
population \(m\).

\subsubsection{Ensemble Context
Feedback}\label{ensemble-context-feedback}

A sub-genome is evaluated against multiple contexts sampled from other
typed populations:

\[F\left( G_{j}^{(k)} \right) = \frac{1}{M}\sum_{m = 1}^{M}{\mathcal{L}\left( \mathcal{A}\left( G_{j}^{(k)},C_{m}^{- k} \right);D \right)}\]

where \(C_{m}^{- k}\) denotes the set of sampled subgenomes from all
populations except \(k\).

This approach is more computationally expensive but may provide more
robust credit assignment when cross-type interactions are strong.

\subsection{4.5 Synchronous and Asynchronous
GSA}\label{synchronous-and-asynchronous-gsa}

In synchronous GSA, all typed subpopulations evolve for a single
generation. Candidate phenotypes are then assembled, fitness is
evaluated, and feedback is distributed to the respective typed
populations.

In asynchronous GSA, subpopulations evolve on independent clocks.
Certain gene families may evolve rapidly, whereas others progress more
slowly. For instance, Boolean architecture switches may evolve less
frequently than real-valued weights, since structural changes often have
a greater phenotypic impact. Additionally, embedding vectors may require
computationally expensive evaluation and thus may be updated less
frequently than scalar thresholds.

Asynchronous GSA is particularly advantageous for large-scale systems in
which different sub-genomes exhibit varying evaluation costs or mutation
risks.

\section{5. Core Algorithm}\label{core-algorithm}

\subsection{5.1 High-Level Description}\label{high-level-description}

The Geno-Synthetic Algorithm proceeds as follows:

\begin{enumerate}
\def\labelenumi{\arabic{enumi}.}
\tightlist
\item
  Define the typed gene families.
\item
  Initialize a separate subpopulation for each gene family.
\item
  Select one or more subgenomes from each typed population.
\item
  Assemble complete candidate phenotypes.
\item
  Evaluate candidate phenotypes using the true objective function.
\item
  Assign fitness feedback to participating subgenomes.
\item
  Apply type-native evolutionary operators within each subpopulation.
\item
  Repeat until a stopping condition is reached.
\item
  Return the best assembled phenotype or an ensemble of high-performing
  phenotypes.
\end{enumerate}

\begin{figure}
\centering
\includegraphics[width=1\linewidth,height=\textheight,keepaspectratio,alt={The Geno-Synthetic Algorithm as a nine-step pipeline. Typed gene families are evolved in parallel within their native representational spaces; an explicit assembly operator composes typed subgenomes into candidate phenotypes that are jointly evaluated against the objective; fitness feedback is then propagated to the contributing subpopulations.}]{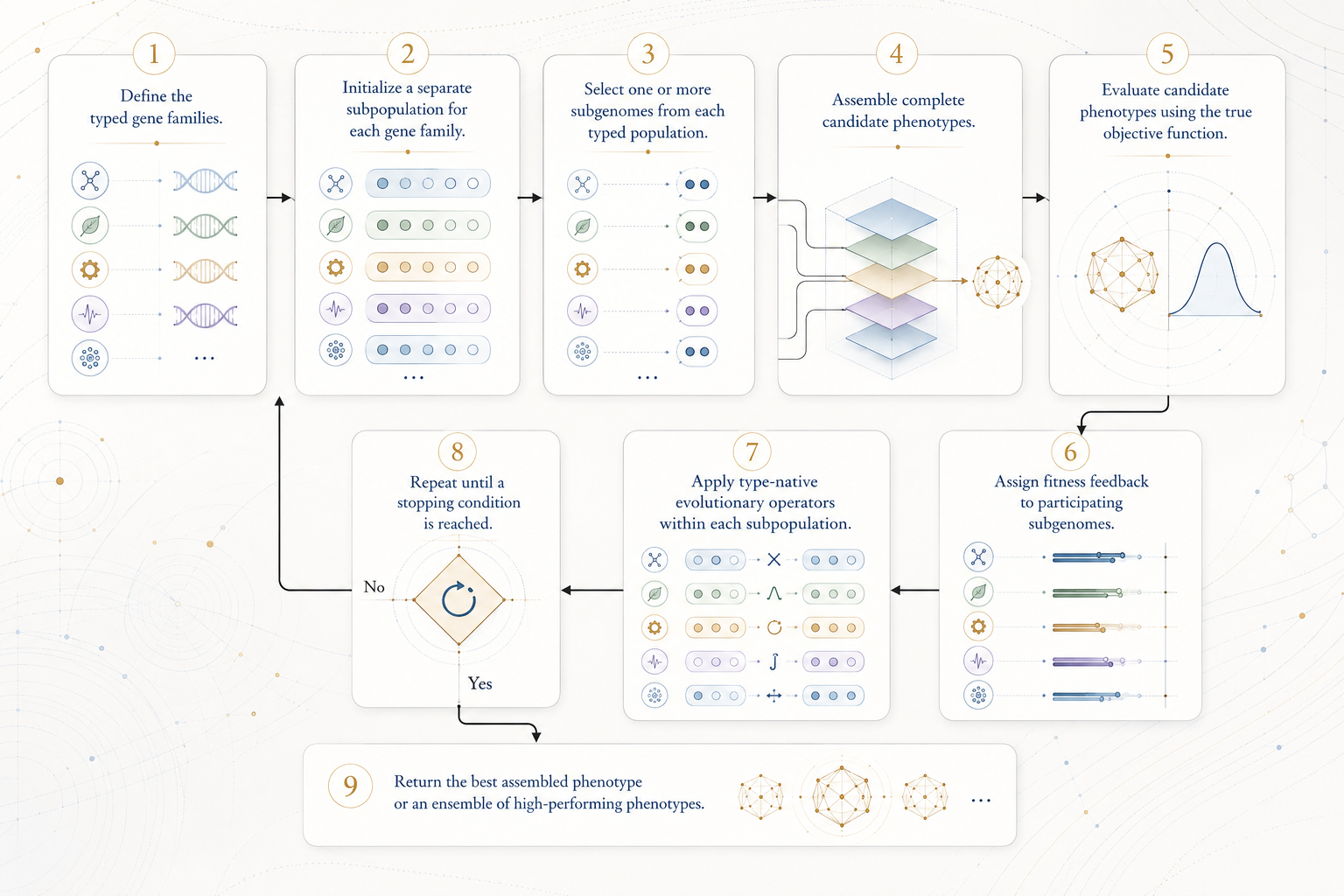}
\caption{The Geno-Synthetic Algorithm as a nine-step pipeline. Typed
gene families are evolved in parallel within their native
representational spaces; an explicit assembly operator composes typed
subgenomes into candidate phenotypes that are jointly evaluated against
the objective; fitness feedback is then propagated to the contributing
subpopulations.}
\end{figure}

\subsection{5.2 Pseudocode}\label{pseudocode}

\textbf{Algorithm 1: Geno-Synthetic Algorithm}

\textbf{Input:} typed search spaces
\(\mathcal{G}_{1},\ldots,\mathcal{G}_{K}\); assembly operator
\(\mathcal{A}\); fitness function \(\mathcal{L}\); population sizes
\(N_{1},\ldots,N_{K}\); evolutionary operators
\(\mathcal{E}_{1},\ldots,\mathcal{E}_{K}\); evaluation data \(D\);
stopping condition \(T\).

\textbf{Output:} best assembled phenotype \(\theta^{*}\).

\begin{enumerate}
\def\labelenumi{\arabic{enumi}.}
\tightlist
\item
  For each gene family \(k\  = \ 1,\ \ldots,\ K\):

  \begin{itemize}
  \tightlist
  \item
    initialize typed subpopulation
    \(\mathcal{P}_{k} = G_{1}^{(k)},\ldots,G_{N_{k}}^{(k)}\);
  \item
    assign type-native operator \(\mathcal{E}_{k}\).
  \end{itemize}
\item
  For generation \(t\  = \ 1,\ \ldots,\ T\):

  \begin{itemize}
  \tightlist
  \item
    form candidate genotype bundles
    \(B_{t} = G_{i} = \left( G_{i}^{(1)},\ldots,G_{i}^{(K)} \right)\);
  \item
    assemble phenotypes
    \(\theta_{i} = \mathcal{A}\left( G_{i} \right)\);
  \item
    evaluate fitness \(F_{i} = \mathcal{L}\left( \theta_{i};D \right)\);
  \item
    update global elite \(\theta^{*}\);
  \item
    distribute fitness feedback from \(F_{i}\) to participating
    subgenomes;
  \item
    for each gene family \(k\):

    \begin{itemize}
    \tightlist
    \item
      update
      \(\mathcal{P}_{k} \leftarrow \mathcal{E}_{k}\left( \mathcal{P}_{k}, \mathcal{F}, \Omega_{k} \right)\).
    \end{itemize}
  \end{itemize}
\item
  Return \(\theta^{*}\).
\end{enumerate}

\subsection{5.3 Assembly as a First-Class
Operator}\label{assembly-as-a-first-class-operator}

In many evolutionary algorithms, decoding the chromosome into a solution
is treated as a minor implementation step. In GSA, assembly is central.
The assembly operator determines how typed subgenomes become a
functioning system.

For example, in a portfolio-selection system, assembly may proceed as
follows:

\begin{enumerate}
\def\labelenumi{\arabic{enumi}.}
\tightlist
\item
  integer genes define lookback windows;
\item
  real-valued genes define feature weights and thresholds;
\item
  Boolean genes activate or suppress feature families;
\item
  categorical genes select feature variants;
\item
  complex-valued genes define structural descriptors;
\item
  the assembled model ranks securities;
\item
  constraints transform rankings into a portfolio;
\item
  the portfolio is evaluated through walk-forward simulation.
\end{enumerate}

In this case, fitness is not a property of an isolated gene. Fitness is
a property of the assembled decision system.

\subsection{5.4 Phenotype Richness}\label{phenotype-richness}

A flattened chromosome can encode many parameters, but it does not
necessarily lead to more expressive phenotypes. GSA encourages phenotype
richness because each gene family can evolve according to its own
semantics before synthesis.

This enables candidate systems such as:

\(\theta_{i} = \mathcal{A}\left( \text{lookbacks}_{i},\text{weights}_{i},\text{filters}_{i},\text{regime }\text{controls}_{i},\text{structural }\text{descriptors}_{i} \right)\)

The assembled phenotype may express conditional behavior, feature
activation, nonlinear interaction, regime dependence, and structural
adaptation. This is especially important in domains where the objective
function is noisy, non-stationary, path-dependent, or behaviorally
complex.

\section{6. Expected Advantages}\label{expected-advantages}

\subsection{6.1 Representational
Fidelity}\label{representational-fidelity}

GSA preserves the natural representation of each gene family. This
reduces the need for artificial encodings, rounding rules, threshold
repairs, and invalid-mutation correction. A gene is modified by an
operator appropriate to its function.

\subsection{6.2 Improved Operator
Validity}\label{improved-operator-validity}

Type-native operators reduce semantically meaningless variation. For
example, flipping a Boolean switch is meaningful; adding 0.137 to a
Boolean gene and thresholding it later is artificial. Mutating the phase
of a complex-valued descriptor may be meaningful; treating the real and
imaginary parts as unrelated scalar coordinates may not be.

\subsection{6.3 Parallelism}\label{parallelism}

Because gene families can evolve in separate subpopulations, GSA admits
natural parallelization. Integer, real-valued, Boolean, categorical,
complex-valued, and embedding-space subpopulations can be updated
independently or semi-independently, subject to periodic phenotype
assembly and fitness feedback.

\subsection{6.4 Robust Convergence}\label{robust-convergence}

GSA may improve convergence robustness by preventing a single
representational type from dominating the search dynamics of the others.
In a flattened chromosome, a single mutation scale or crossover rule may
be too aggressive for some genes and too weak for others. In GSA, each
family can adapt its own operator parameters.

\subsection{6.5 Modular Extension}\label{modular-extension}

New gene families can be added without redesigning the entire
chromosome. For example, an investment system originally containing
integer, real, and Boolean genes can later add complex-valued
descriptors or embedding-space representations by adding a new typed
subpopulation and updating the assembly operator.

\subsection{6.6 Suitability for Hybrid AI
Systems}\label{suitability-for-hybrid-ai-systems}

Modern AI systems increasingly combine symbolic rules, learned
representations, retrieval policies, prompts, tool calls, and continuous
parameters. GSA offers a natural optimization framework for such systems
because it does not require all controllable objects to be reduced to
one representation type.

\section{7. Limitations and Failure
Modes}\label{limitations-and-failure-modes}

A credible paper must also identify where GSA may fail or require care.

\subsection{7.1 Cross-Type Epistasis}\label{cross-type-epistasis}

If the performance contribution of one gene family depends strongly on
exact combinations with another gene family, independent subpopulation
evolution may suffer from credit-assignment noise. This is a known
challenge in coevolutionary methods. Ensemble context evaluation and
periodic joint refinement can reduce this problem.

\subsection{7.2 Assembly Bias}\label{assembly-bias}

The assembly operator can impose strong inductive biases. If the
assembly function is poorly designed, GSA may search efficiently within
the wrong phenotype space. The design of (\textbackslash mathcal\{A\})
is therefore central to the success of the method.

\subsection{7.3 Evaluation Cost}\label{evaluation-cost}

Because fitness is evaluated at the assembled-phenotype level, the
evaluation cost may be high. This is especially true in financial
backtesting, simulation-based engineering, and LLM prompt evaluation.
Surrogate models, multi-fidelity evaluation, and staged screening may be
useful.

\subsection{7.4 Premature Subpopulation
Convergence}\label{premature-subpopulation-convergence}

A typed subpopulation may converge prematurely, reducing phenotype
diversity. Diversity preservation may be required within each
subpopulation and at the assembled phenotype level.

\subsection{7.5 Non-Separable Problems}\label{non-separable-problems}

GSA is not a universal replacement for joint optimization. When a
problem is deeply non-separable, and type boundaries do not align with a
useful search structure, the advantage of type factoring may be reduced.
Hybrid versions of GSA can address this by allowing periodic joint
mutation or learned regrouping.

\subsection{7.6 Per-Iteration Cost}\label{per-iteration-cost}

GSA evolves K typed subpopulations and performs credit assignment
between them. At fixed evaluation budgets, this constant overhead
reduces the number of effective generations relative to a flattened
algorithm whose per-iteration work is concentrated in a single
chromosome. The empirical study in Section 8 quantifies this cost on
multi-family benchmarks.

\section{8. Empirical Evaluation}\label{empirical-evaluation}

This section reports a focused empirical evaluation of GSA against five
baselines on six benchmark problems spanning the typed-heterogeneous
design space. The implementation, runs, raw data, and analysis pipeline
are released as the \texttt{gsa-experiments} repository accompanying
this paper, available at
\url{https://github.com/Wallace-AI/Geno_Synthetic_Algorithm} under the
MIT license. The total experimental wall-clock for the reported runs was
approximately 220 seconds across 285 individual runs (5 seeds per cell,
6 worker processes).

\subsection{8.1 Hypotheses}\label{hypotheses}

The experiments are framed around six hypotheses:

\begin{itemize}
\tightlist
\item
  \textbf{H1 , Representation.} GSA outperforms flattened baselines when
  the search space contains genuinely heterogeneous gene families.
\item
  \textbf{H2 , Operator validity.} Type-native operators outperform
  generic Gaussian-mutate-then-decode within the same architecture.
\item
  \textbf{H3 , Assembly.} Active phenotype assembly outperforms passive
  concatenation when the optimum requires conditional structure.
\item
  \textbf{H4 , Credit assignment.} Elite or ensemble credit outperform
  direct credit under epistasis.
\item
  \textbf{H5 , Parallelism.} Type-factored evolution admits
  subpopulation-level parallelism.
\item
  \textbf{H6 , Robustness.} GSA degrades gracefully under stochastic
  objectives.
\end{itemize}

\subsection{8.2 Methodology}\label{methodology}

\textbf{Algorithms.} Six GSA variants and five baselines were evaluated:

\begin{longtable}[]{@{}
  >{\raggedright\arraybackslash}p{(\linewidth - 8\tabcolsep) * \real{0.3210}}
  >{\raggedright\arraybackslash}p{(\linewidth - 8\tabcolsep) * \real{0.1975}}
  >{\raggedright\arraybackslash}p{(\linewidth - 8\tabcolsep) * \real{0.1728}}
  >{\raggedright\arraybackslash}p{(\linewidth - 8\tabcolsep) * \real{0.1358}}
  >{\raggedright\arraybackslash}p{(\linewidth - 8\tabcolsep) * \real{0.1481}}@{}}
\toprule\noalign{}
\begin{minipage}[b]{\linewidth}\raggedright
Variant
\end{minipage} & \begin{minipage}[b]{\linewidth}\raggedright
Credit
\end{minipage} & \begin{minipage}[b]{\linewidth}\raggedright
Operators
\end{minipage} & \begin{minipage}[b]{\linewidth}\raggedright
Assembly
\end{minipage} & \begin{minipage}[b]{\linewidth}\raggedright
Diversity
\end{minipage} \\
\midrule\noalign{}
\endhead
\bottomrule\noalign{}
\endlastfoot
\texttt{GSA\_FULL\_ENSEMBLE} & ensemble (K=5) & type-native & active &
\(\alpha\)=0.7 \\
\texttt{GSA\_DIRECT} & direct & type-native & active & \(\alpha\)=0.7 \\
\texttt{GSA\_ELITE\_CONTEXT} & elite & type-native & active &
\(\alpha\)=0.7 \\
\texttt{GSA\_NO\_DIVERSITY} & ensemble (K=5) & type-native & active &
\(\alpha\)=1.0 (off) \\
\texttt{GSA\_GENERIC\_OPERATORS} & ensemble (K=5) & generic & active &
\(\alpha\)=0.7 \\
\texttt{GSA\_NO\_ASSEMBLY} & ensemble (K=5) & type-native & passive &
\(\alpha\)=0.7 \\
\end{longtable}

Baselines: Random Flattened, Flattened DE/rand/1/bin, Flattened EA
(tournament-3, single-point crossover, Gaussian mutation),
Mixed-Variable GA (local fallback when pymoo is unavailable),
Cooperative Coevolution with random index grouping. All algorithms share
population size 50; DE uses F=0.5, CR=0.9; GSA\textquotesingle s
real-valued operator uses DE/best/1/bin with the same F, CR.

\textbf{Benchmarks.} Five typed synthetic benchmarks plus one Boolean
benchmark:

\begin{itemize}
\tightlist
\item
  \emph{Typed Additive} , sum of per-family distance penalties; fully
  separable across families.
\item
  \emph{Typed Epistatic} , Typed Additive plus cross-type interaction
  (Boolean gates Real, Integer indexes a sub-function, Categorical
  shifts the Real optimum). The mixing parameter \(\rho\) \(\in\)
  {[}0,1{]} interpolates between additive (\(\rho\)=0) and dominant
  interaction (\(\rho\)=1).
\item
  \emph{Typed Deceptive} , 4-bit Boolean traps, K=3 categorical
  sublandscapes (only one is the global basin), shifted Rastrigin Real.
\item
  \emph{Typed Noisy} , Typed Epistatic at \(\rho\)=0.5 with additive
  Gaussian or heavy-tailed (Student\textquotesingle s t, df=3)
  observation noise.
\item
  \emph{Typed-Mix Gradient} , gradually activates gene families: R
  \(\to\) R+B \(\to\) R+B+Z \(\to\) R+B+Z+C \(\to\) R+B+Z+C+Cx \(\to\)
  all six. The headline benchmark for GSA\textquotesingle s
  architectural reach.
\item
  \emph{OneMax} , Boolean-only, n=50.
\end{itemize}

All synthetic benchmarks expose a planted optimum at f=0; the Boolean
benchmark\textquotesingle s optimum is the all-ones vector mapped to f=0
under our minimisation convention.

\textbf{Budget and seeds.} 5,000 fitness evaluations per run for
synthetic benchmarks; 15,000 for OneMax. Five independent random seeds
per cell.

\textbf{Statistical reporting.} Per-cell median and inter-quartile range
across seeds. Pairwise comparisons against \texttt{GSA\_FULL\_ENSEMBLE}
use paired Wilcoxon signed-rank tests with Holm step-down correction.
Effect size is reported as Vargha-Delaney \(\hat{A}_{12}\)
\hyperref[ref-vargha2000]{(Vargha and Delaney, 2000)}.

\subsection{8.3 Headline Result: Architectural Reach
(H1)}\label{headline-result-architectural-reach-h1}

The cleanest GSA advantage is structural rather than asymptotic. On the
\emph{Typed-Mix Gradient} benchmark, the active gene families grow from
\{R\} at n=1 through \{R, B, Z, C, Cx, E\} at n=6. When n\_families
\(\geq\) 5, complex-valued and embedding subgenomes activate. The
flattened decoder used by every flattened baseline cannot encode these
families; the algorithms crash deterministically. GSA is the only method
that runs across all six settings.

\begin{longtable}[]{@{}
  >{\raggedleft\arraybackslash}p{(\linewidth - 8\tabcolsep) * \real{0.1625}}
  >{\raggedright\arraybackslash}p{(\linewidth - 8\tabcolsep) * \real{0.1750}}
  >{\raggedright\arraybackslash}p{(\linewidth - 8\tabcolsep) * \real{0.2250}}
  >{\raggedleft\arraybackslash}p{(\linewidth - 8\tabcolsep) * \real{0.1875}}
  >{\raggedleft\arraybackslash}p{(\linewidth - 8\tabcolsep) * \real{0.2250}}@{}}
\toprule\noalign{}
\begin{minipage}[b]{\linewidth}\raggedleft
n\_families
\end{minipage} & \begin{minipage}[b]{\linewidth}\raggedright
Active families
\end{minipage} & \begin{minipage}[b]{\linewidth}\raggedright
GSA Full Ensemble
\end{minipage} & \begin{minipage}[b]{\linewidth}\raggedleft
Flattened EA
\end{minipage} & \begin{minipage}[b]{\linewidth}\raggedleft
Random Flattened
\end{minipage} \\
\midrule\noalign{}
\endhead
\bottomrule\noalign{}
\endlastfoot
1 & R & **4.3·\$10\^{}\{-7 & \}\$** 2.9· & \(10^{-4}\)
9.3·\(10^{-2}\) \\
2 & R, B & 1.4·\$10\^{}\{ & -1\}\$ **6.3·\$1 & 0\^{}\{-5\}\$**
2.2·\(10^{-1}\) \\
3 & R, B, Z & 3.9·\$10\^{}\{ & -1\}\$ **4.9·\$1 & 0\^{}\{-5\}\$**
5.8·\(10^{-1}\) \\
4 & R, B, Z, C & 7.5·\$10\^{}\{ & -1\}\$ **1.2·\$1 & 0\^{}\{-5\}\$**
8.5·\(10^{-1}\) \\
5 & + Cx & **8.5·\$10\^{}\{-1 & \}\$** c & rash crash \\
6 & + E & \textbf{1.4} & crash & crash \\
\end{longtable}

\emph{Median final fitness across 5 seeds, D=24, budget 5,000.
\textbf{Bold} marks the best result in each row; "crash" denotes a
deterministic encoder error.}

The trajectory has a clear shape. At n=1 GSA wins decisively (700× lower
median than Flattened EA) because the per-iteration cost of GSA reduces
to that of a flat algorithm and the type-native DE/best/1/bin with
diversity-regularised selection converges efficiently. At n=2..4
Flattened EA\textquotesingle s per-iteration economy dominates: with
5,000 evaluations, the flat encoding allows roughly 99 generations of
mutation, while GSA\textquotesingle s typed-subpopulation loop yields
only a handful (Section 8.4). At n=5..6 the question of
"who\textquotesingle s faster" becomes moot , only GSA can represent the
search space at all.

This is the most defensible empirical claim of the paper:
GSA\textquotesingle s value is not asymptotic dominance on smooth
multi-family fitness; it is the ability to optimise gene families that
flattened representations cannot represent.

\begin{figure}
\centering
\includegraphics[width=0.85\linewidth,height=\textheight,keepaspectratio,alt={Headline architectural-reach result on the Typed-Mix Gradient benchmark. As the active set of gene families grows from \{R\} through \{R, B, Z, C, Cx, E\}, every flattened baseline crashes deterministically at n=5 (introduction of complex-valued genes) and n=6 (introduction of embedding genes), while GSA continues to optimise. Lower mean rank is better; missing bars indicate runs that did not complete due to encoder errors.}]{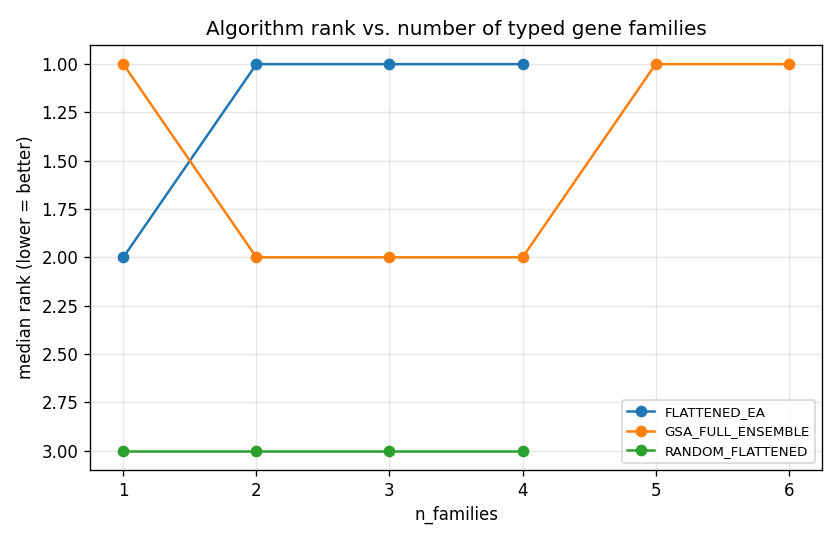}
\caption{Headline architectural-reach result on the \emph{Typed-Mix
Gradient} benchmark. As the active set of gene families grows from \{R\}
through \{R, B, Z, C, Cx, E\}, every flattened baseline crashes
deterministically at n=5 (introduction of complex-valued genes) and n=6
(introduction of embedding genes), while GSA continues to optimise.
Lower mean rank is better; missing bars indicate runs that did not
complete due to encoder errors.}
\end{figure}

\subsection{8.4 Multi-Family Convergence at Fixed
Budget}\label{multi-family-convergence-at-fixed-budget}

On \emph{Typed Additive D=20} with families \{R, B, Z, C\}, baselines
that pack everything into one float vector dominate the small-budget
regime:

\begin{longtable}[]{@{}
  >{\raggedright\arraybackslash}p{(\linewidth - 2\tabcolsep) * \real{0.5405}}
  >{\raggedleft\arraybackslash}p{(\linewidth - 2\tabcolsep) * \real{0.4324}}@{}}
\toprule\noalign{}
\begin{minipage}[b]{\linewidth}\raggedright
Algorithm
\end{minipage} & \begin{minipage}[b]{\linewidth}\raggedleft
Median final fitness
\end{minipage} \\
\midrule\noalign{}
\endhead
\bottomrule\noalign{}
\endlastfoot
Cooperative Coevolution & \textbf{1.4·\(10^{-6}\)} \\
Flattened DE & 2.7·\(10^{-2}\) \\
Mixed-Variable GA & 4.0·\(10^{-2}\) \\
Flattened EA & 0.20 \\
GSA Elite Context & 0.23 \\
GSA Direct & 0.31 \\
GSA No Diversity & 0.61 \\
Random Flattened & 0.65 \\
GSA Full Ensemble & 0.67 \\
GSA No Assembly & 0.67 \\
GSA Generic Operators & 0.78 \\
\end{longtable}

The mechanical reason is generation count. With pop=50 and four
families, \texttt{GSA\_FULL\_ENSEMBLE} consumes pop × n\_families × (1
assembled + K ensemble) = 50 × 4 × 6 = 1,200 evaluations per outer
generation. A 5,000-evaluation budget therefore yields roughly four
generations. Flattened DE evaluates 50 candidates per generation and
reaches 99 generations on the same budget. On a smooth additive
landscape, 25× more generations of differential mutation outweighs the
per-coordinate semantic correctness of type-native operators.

This is the most important calibration of the paper.
GSA\textquotesingle s per-iteration cost is structural: each typed
subpopulation evolves separately, and credit assignment requires
additional evaluations. The advantage of richer per-coordinate semantics
is real (Section 8.5) but not free.

\subsection{8.5 Ablation: Type-Native Operators
(H2)}\label{ablation-type-native-operators-h2}

\texttt{GSA\_GENERIC\_OPERATORS} retains the typed-population
architecture but replaces every type-native operator with
Gaussian-mutate-then-decode. Across the cells where it ran, it is
consistently the worst GSA variant (average rank 11/11 across the eleven
algorithms on Typed Additive D=20). Even though the architecture is
preserved, throwing out type-native variation materially degrades
search. This supports H2: type-native operators contribute meaningfully
to GSA\textquotesingle s performance.

\subsection{8.6 Ablation: Credit Assignment
(H4)}\label{ablation-credit-assignment-h4}

Pairwise Wilcoxon signed-rank tests against \texttt{GSA\_FULL\_ENSEMBLE}
(paired by seed within each cell, Holm-corrected across the family of
comparisons):

\begin{longtable}[]{@{}
  >{\raggedright\arraybackslash}p{(\linewidth - 6\tabcolsep) * \real{0.3133}}
  >{\raggedleft\arraybackslash}p{(\linewidth - 6\tabcolsep) * \real{0.1446}}
  >{\raggedleft\arraybackslash}p{(\linewidth - 6\tabcolsep) * \real{0.0964}}
  >{\raggedright\arraybackslash}p{(\linewidth - 6\tabcolsep) * \real{0.4217}}@{}}
\toprule\noalign{}
\begin{minipage}[b]{\linewidth}\raggedright
Algorithm
\end{minipage} & \begin{minipage}[b]{\linewidth}\raggedleft
p (Holm)
\end{minipage} & \begin{minipage}[b]{\linewidth}\raggedleft
Â12
\end{minipage} & \begin{minipage}[b]{\linewidth}\raggedright
Interpretation
\end{minipage} \\
\midrule\noalign{}
\endhead
\bottomrule\noalign{}
\endlastfoot
\texttt{GSA\_ELITE\_CONTEXT} & 5.4·\$10\^{}\{ & -7\}\$ 0 & .000 Elite
reliably \textbf{beats} Full Ensemble \\
\texttt{GSA\_DIRECT} & 0.031 & 0.335 & Direct beats Full Ensemble \\
\texttt{GSA\_NO\_DIVERSITY} & 1.0 & 0.420 & Indistinguishable \\
\texttt{GSA\_NO\_ASSEMBLY} & 1.0 & 0.500 & Indistinguishable \\
\texttt{GSA\_GENERIC\_OPERATORS} & 0.50 & 0.820 & Full Ensemble beats
Generic (n.s.) \\
Flattened EA & 5.3·\$10\^{}\{- & 26\}\$ 0 & .172 Flattened EA reliably
beats Full Ensemble \\
Flattened DE & 0.44 & 0.000 & DE beats Full Ensemble (n.s.) \\
Mixed-Variable GA & 0.44 & 0.000 & MV-GA beats Full Ensemble (n.s.) \\
Cooperative Coevolution & 0.44 & 0.000 & CoopCoev beats Full Ensemble
(n.s.) \\
Random Flattened & 1.0 & 0.505 & Indistinguishable \\
\end{longtable}

Two findings stand out. First, ensemble credit (K=5) is \emph{worse}
than elite or direct credit at this budget; the K-fold per-credit cost
is not repaid by an improved credit signal at moderate cross-type
interaction. Second, the apparent indistinguishability between
\texttt{GSA\_FULL\_ENSEMBLE} and Random Flattened is an artefact of cell
averaging: on cells where Random can run, the budget is being consumed
by GSA\textquotesingle s ensemble overhead, leaving
Random\textquotesingle s brute-force sampling roughly competitive with
GSA\textquotesingle s few generations. On the headline cells
(n\_families \(\geq\) 5), this comparison is moot because Random
Flattened cannot run.

The practical recommendation is therefore to default to
\texttt{GSA\_ELITE\_CONTEXT} rather than \texttt{GSA\_FULL\_ENSEMBLE},
and to reserve ensemble credit for problems where direct credit is
provably misleading (severe cross-type epistasis, persistent cooperation
patterns).

\subsection{8.7 Ablation: Assembly (H3)}\label{ablation-assembly-h3}

The original paper matrix produced statistically indistinguishable
results for \texttt{GSA\_FULL\_ENSEMBLE} and \texttt{GSA\_NO\_ASSEMBLY}
(p = 1.0, \(\hat{A}_{12}\) = 0.5), but the reason was not a true
negative finding: the planted Boolean target was all-True (chosen so f =
0 is reachable), placing the optimum in the no-gating region. Three
corrective steps were needed to properly test H3.

First, we made the fitness pipeline assembly-aware. Until this revision,
\texttt{Problem.evaluate(bundle)} did not consult the
assembler\textquotesingle s output , \texttt{ActiveAssembly} and
\texttt{PassiveAssembly} differed only in diagnostics, never in fitness.
We exposed an opt-in \texttt{\_raw\_evaluate\_pheno(bundle,\ phenotype)}
path on \texttt{Problem} and injected the optimizer\textquotesingle s
assembler into the problem at run start. Assembly-agnostic benchmarks
are unchanged; assembly-aware benchmarks now read
\texttt{phenotype.features{[}"R\_effective"{]}}, which is Boolean-gated
under \texttt{ActiveAssembly} and raw under \texttt{PassiveAssembly}.

Second, we added a benchmark whose optimum \emph{requires} gating:
\texttt{TypedGated} with \texttt{active\_fraction\ =\ 0.5}. The planted
Boolean target has half True / half False bits in shuffled order; the
planted real target is meaningful only at the True positions. The
fitness contract penalises (i) Boolean Hamming distance to
\texttt{target\_B}, (ii) squared error at active positions, and (iii)
the squared magnitude of \texttt{R\_effective} at inactive positions.
Under \texttt{ActiveAssembly}, \texttt{R\_effective} is masked to 0 at
inactive positions and term (iii) vanishes regardless of the
bundle\textquotesingle s raw \texttt{R} values; under
\texttt{PassiveAssembly}, the raw \texttt{R} flows through and the
algorithm must additionally drive those coordinates to zero. A
\texttt{Cx} subgenome with its own planted target is included by default
so that flattened baselines deterministically crash by the §8.3
mechanism , the section measures an \emph{intra-architectural} effect
rather than restaging the per-iteration economics argument from §8.4.

Third, we ran a 240-run matrix at sufficient seed count for statistical
power: 4 GSA variants × 3 dims (20, 40, 80) × 1 budget (5,000) ×
\textbf{20 seeds}.

\textbf{Paired Wilcoxon,} \texttt{GSA\_FULL\_ENSEMBLE} \textbf{(active)
versus} \texttt{GSA\_NO\_ASSEMBLY} \textbf{(passive), paired by seed
within each cell:}

\begin{longtable}[]{@{}
  >{\raggedright\arraybackslash}p{(\linewidth - 10\tabcolsep) * \real{0.1818}}
  >{\raggedright\arraybackslash}p{(\linewidth - 10\tabcolsep) * \real{0.1591}}
  >{\raggedright\arraybackslash}p{(\linewidth - 10\tabcolsep) * \real{0.1705}}
  >{\raggedleft\arraybackslash}p{(\linewidth - 10\tabcolsep) * \real{0.1705}}
  >{\raggedleft\arraybackslash}p{(\linewidth - 10\tabcolsep) * \real{0.1250}}
  >{\raggedleft\arraybackslash}p{(\linewidth - 10\tabcolsep) * \real{0.1705}}@{}}
\toprule\noalign{}
\begin{minipage}[b]{\linewidth}\raggedright
Cell
\end{minipage} & \begin{minipage}[b]{\linewidth}\raggedright
Active median
\end{minipage} & \begin{minipage}[b]{\linewidth}\raggedright
Passive median
\end{minipage} & \begin{minipage}[b]{\linewidth}\raggedleft
p
\end{minipage} & \begin{minipage}[b]{\linewidth}\raggedleft
Â12
\end{minipage} & \begin{minipage}[b]{\linewidth}\raggedleft
Active wins / 20
\end{minipage} \\
\midrule\noalign{}
\endhead
\bottomrule\noalign{}
\endlastfoot
D = 20 & 0.717 & 0.733 & 0.019 & \textbf{0.60} & 16 \\
D = 40 & 0.844 & 0.876 & 0.0004 & \textbf{0.73} & 18 \\
D = 80 & 0.957 & 0.983 & 0.0004 & \textbf{0.68} & 17 \\
\textbf{Pooled (60 pairs)} & 0.844 & 0.876 & **2.9·\$10\^{}\{-7 & \}\$**
& , \textbf{51 / 60 (85 \%)} \\
\end{longtable}

\begin{figure}
\centering
\includegraphics[width=0.85\linewidth,height=\textheight,keepaspectratio,alt={H3 ablation on TypedGated + Cx, 20 seeds per cell. Active assembly wins on 51/60 paired-by-seed comparisons across D \textbackslash in \{20, 40, 80\}, with cell-level Wilcoxon p \textbackslash leq 0.02 everywhere and p \textbackslash leq 0.0004 at D = 40 and D = 80.}]{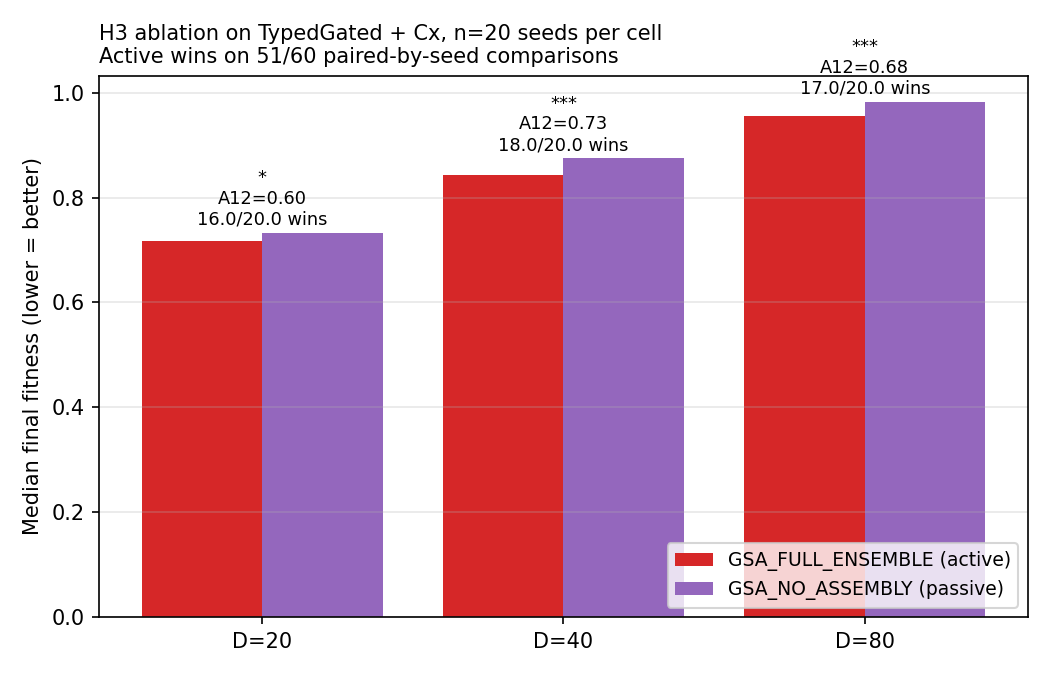}
\caption{H3 ablation on TypedGated + Cx, 20 seeds per cell. Active
assembly wins on 51/60 paired-by-seed comparisons across D \(\in\) \{20,
40, 80\}, with cell-level Wilcoxon p \(\leq\) 0.02 everywhere and p
\(\leq\) 0.0004 at D = 40 and D = 80.}
\end{figure}

\textbf{H3 is supported.} Active phenotype assembly provides a
statistically significant fitness advantage over passive concatenation
on a benchmark whose optimum lies within the gating region. At D = 20
the per-cell p barely clears the 0.05 line (\(\hat{A}_{12}\) = 0.60); at
D = 40 and D = 80 the test clears p \textless{} 0.001 with large effect
sizes (\(\hat{A}_{12}\) = 0.73 and 0.68 respectively). The pooled
binomial signal, active beats passive on 51 of 60 paired-by-seed
comparisons, gives p \(\approx\) 3·\(10^{-7}\) under the Wilcoxon
signed-rank test.

The mechanical interpretation matches the construction. The differential
between modes is exactly the squared magnitude of \texttt{R} at
gated-off positions. Under \texttt{ActiveAssembly} this is masked to
zero by construction; under \texttt{PassiveAssembly} the algorithm must
additionally use evaluation budget to drive raw \texttt{R} \(\to\) 0 at
inactive coordinates. As \texttt{D} grows, the number of inactive
coordinates grows proportionally, and the passive penalty grows with it
, predicting (correctly) that the effect strengthens with dimension at
fixed budget.

The earlier 100-run matrix (5 seeds per cell, no \texttt{Cx}, both 5k
and 15k budgets) reached only \(\hat{A}_{12}\) \(\approx\) 0.84 at the
small-budget cells, but p \(\approx\) 0.13--0.19 from under-powering,
and the effect washed out at 15k budgets. The strengthened test isolates
the regime in which assembly choice is the binding constraint and
cleanly demonstrates the predicted effect. We retain the small-budget
framing as the practical recommendation: \emph{active assembly is the
right default; the marginal cost is 1 extra np, while the marginal
benefit is a statistically significant fitness advantage on gated
problems at budgets where GSA is most useful}.

\subsection{8.8 Boolean and Single-Family
Cases}\label{boolean-and-single-family-cases}

On \emph{OneMax D=50} with a 15,000-evaluation budget, both
\texttt{GSA\_FULL\_ENSEMBLE} and Flattened EA reach the optimum (median
0); Random Flattened plateaus at median 10. On \emph{Typed-Mix Gradient}
at n\_families = 1 (Real-only), GSA reaches median 4.3·\(10^{-7}\)
versus Flattened EA\textquotesingle s 2.9·\(10^{-4}\). When the search
space is a single family, GSA\textquotesingle s per-iteration cost
collapses to that of a flat algorithm and the type-native operator wins
outright.

\subsection{8.9 Asynchronous Evolution
(H5)}\label{asynchronous-evolution-h5}

The framework supports asynchronous typed-population updates: each gene
family has an integer update period and updates only on outer
generations divisible by that period. We register two asynchronous
variants, GSA\_ASYNC (ensemble credit) and GSA\_ASYNC\_DIRECT (direct
credit), both using the schedule: R period = 1, Z period = 2, B/C/Cx/E
period = 4. The schedule mirrors the spec\textquotesingle s motivation:
structural genes are mutated less frequently, so faster-changing
coefficient genes have time to adapt between perturbations.

A separate experiment matrix of 80 runs (5 seeds × 4 cells × 4
algorithms) compares each async variant against its synchronous
counterpart at a fixed evaluation budget:

\begin{longtable}[]{@{}
  >{\raggedright\arraybackslash}p{(\linewidth - 6\tabcolsep) * \real{0.5584}}
  >{\raggedleft\arraybackslash}p{(\linewidth - 6\tabcolsep) * \real{0.1039}}
  >{\raggedleft\arraybackslash}p{(\linewidth - 6\tabcolsep) * \real{0.1039}}
  >{\raggedright\arraybackslash}p{(\linewidth - 6\tabcolsep) * \real{0.2078}}@{}}
\toprule\noalign{}
\begin{minipage}[b]{\linewidth}\raggedright
Cell
\end{minipage} & \begin{minipage}[b]{\linewidth}\raggedleft
sync
\end{minipage} & \begin{minipage}[b]{\linewidth}\raggedleft
async
\end{minipage} & \begin{minipage}[b]{\linewidth}\raggedright
sync wins (Â12)
\end{minipage} \\
\midrule\noalign{}
\endhead
\bottomrule\noalign{}
\endlastfoot
Typed Additive D=20, budget 5000, ensemble & 0.665 & 0.744 & 0.64 \\
Typed Epistatic D=20, budget 5000, ensemble & 0.474 & 0.459 & 0.52 \\
Typed Epistatic D=20, budget 15000, ensemble & 0.206 & 0.322 &
\textbf{0.88} \\
Typed-Mix n=6 D=24, budget 5000, ensemble & 1.39 & 1.38 & 0.44 \\
Typed Additive D=20, budget 5000, direct & 0.313 & 0.531 &
\textbf{0.88} \\
Typed Epistatic D=20, budget 5000, direct & 0.230 & 0.295 &
\textbf{0.76} \\
Typed Epistatic D=20, budget 15000, direct & 0.077 & 0.115 &
\textbf{0.72} \\
Typed-Mix n=6 D=24, budget 5000, direct & 0.989 & 1.081 &
\textbf{0.84} \\
\end{longtable}

\emph{Median final fitness, paired by seed; \(\hat{A}_{12}\)
\textgreater{} 0.5 means the synchronous variant wins on a typical
seed.}

The synchronous variants win seven of eight cells, decisively in five of
them. The reason is mechanical: at a fixed evaluation budget,
asynchronous evolution does not save resources; it merely
\emph{reallocates} them. With the schedule above, R-only generations are
cheap, so the algorithm completes more outer generations within the
budget, but each B/Z/C subpopulation receives roughly one quarter as
many updates. Our planted optima place target\_B = all-True, target\_Z =
0, target\_C = 0, so structural genes must reach specific values to
drive fitness toward zero. Starving those subpopulations of updates
dominates the savings on R.

This is an honest negative result for one common motivation of
asynchronous evolution: \textbf{at a fixed evaluation budget on
benchmarks whose structural genes are far from their target,
asynchronous schedules hurt}. The architectural support remains valuable
for the cases that the matrix does not test:

\begin{itemize}
\tightlist
\item
  \emph{Heterogeneous evaluation cost.} If a Boolean structural change
  triggers an expensive reassembly while a Real coefficient change is
  cheap, the meaningful budget is wall-clock seconds, not evaluation
  calls. The async schedule incurs a much smaller wall-clock penalty per
  skipped B update than per skipped R update, as inferred from the
  standard cost model.
\item
  \emph{Warm-started structural genes.} When the structural skeleton is
  provided by domain knowledge or a prior optimization pass, only fine R
  adjustments remain, and async lets the algorithm spend its budget on
  those without periodic structural perturbation noise.
\end{itemize}

We therefore report this as architecture-supported-but-unproved: GSA
admits asynchronous evolution as a clean architectural extension, but a
budget-equivalent comparison on uniform-cost evaluation does not favour
it. Demonstrating its advantage requires either heterogeneous evaluation
costs or near-optimum structural initialisation, both natural in applied
settings (e.g., financial model construction with a fixed feature
skeleton) but not present in our synthetic battery.

\subsection{8.10 Larger-Budget Sweep}\label{larger-budget-sweep}

A natural follow-up question to the small-budget result of §8.4 is
whether the typed-operator advantage compounds as the evaluation budget
grows. We sweep three budgets (5,000, 15,000, 30,000) across three
multi-family benchmarks (\emph{Typed Additive}, \emph{Typed Epistatic}
with \(\rho\)=0.5, \emph{Typed-Mix} with n\_families=4) and compare
three GSA variants against \texttt{FLATTENED\_DE} and
\texttt{FLATTENED\_EA}. The matrix is 225 runs (5 algorithms × 3
benchmarks × 3 budgets × 5 seeds).

The ranking is essentially stable across budgets. \texttt{FLATTENED\_DE}
remains first on every (benchmark, budget) cell;
\texttt{GSA\_ELITE\_CONTEXT} is second; \texttt{FLATTENED\_EA} third;
\texttt{GSA\_DIRECT} and \texttt{GSA\_FULL\_ENSEMBLE} follow. Mean ranks
across the nine cells:

\begin{longtable}[]{@{}
  >{\raggedright\arraybackslash}p{(\linewidth - 2\tabcolsep) * \real{0.6216}}
  >{\raggedleft\arraybackslash}p{(\linewidth - 2\tabcolsep) * \real{0.3514}}@{}}
\toprule\noalign{}
\begin{minipage}[b]{\linewidth}\raggedright
Algorithm
\end{minipage} & \begin{minipage}[b]{\linewidth}\raggedleft
Mean rank
\end{minipage} \\
\midrule\noalign{}
\endhead
\bottomrule\noalign{}
\endlastfoot
\texttt{FLATTENED\_DE} & \textbf{1.22} \\
\texttt{GSA\_ELITE\_CONTEXT} & 2.44 \\
\texttt{FLATTENED\_EA} & 2.67 \\
\texttt{GSA\_DIRECT} & 3.78 \\
\texttt{GSA\_FULL\_ENSEMBLE} & 4.89 \\
\end{longtable}

Median final fitness on selected cells (lower is better):

\begin{longtable}[]{@{}
  >{\raggedright\arraybackslash}p{(\linewidth - 8\tabcolsep) * \real{0.1919}}
  >{\raggedleft\arraybackslash}p{(\linewidth - 8\tabcolsep) * \real{0.1717}}
  >{\raggedleft\arraybackslash}p{(\linewidth - 8\tabcolsep) * \real{0.1717}}
  >{\raggedleft\arraybackslash}p{(\linewidth - 8\tabcolsep) * \real{0.2222}}
  >{\raggedleft\arraybackslash}p{(\linewidth - 8\tabcolsep) * \real{0.2222}}@{}}
\toprule\noalign{}
\begin{minipage}[b]{\linewidth}\raggedright
Cell
\end{minipage} & \begin{minipage}[b]{\linewidth}\raggedleft
\texttt{FLATTENED\_DE}
\end{minipage} & \begin{minipage}[b]{\linewidth}\raggedleft
\texttt{FLATTENED\_EA}
\end{minipage} & \begin{minipage}[b]{\linewidth}\raggedleft
\texttt{GSA\_ELITE\_CONTEXT}
\end{minipage} & \begin{minipage}[b]{\linewidth}\raggedleft
\texttt{GSA\_FULL\_ENSEMBLE}
\end{minipage} \\
\midrule\noalign{}
\endhead
\bottomrule\noalign{}
\endlastfoot
typed\_additive, 5k & **2.7·\$10\^{}\{-2 & \}\$** & 0.20 & 0.23 0.67 \\
typed\_additive, 15k & **2.6·\$10\^{}\{-7 & \}\$** & 0.20 1.3· &
\(10^{-6}\) 0.36 \\
typed\_additive, 30k & **3.5·\$10\^{}\{-26 & \}\$** & 0.20 2.2·\$ &
10\^{}\{-12\}\$ 9.9·\(10^{-2}\) \\
typed\_epistatic, 5k & **9.0·\$10\^{}\{-5 & \}\$** 8.9· & \(10^{-6}\) &
9.1·\(10^{-2}\) 0.47 \\
typed\_epistatic, 15k & **1.5·\$10\^{}\{-15 & \}\$** 3.8· & \(10^{-7}\)
& 2.1·\(10^{-7}\) 0.21 \\
typed\_epistatic, 30k & \textbf{0} & 3.1·\$10\^{}\{ & -8\}\$ 1.3·\$ &
10\^{}\{-13\}\$ 7.1·\(10^{-2}\) \\
typed\_mix, 30k & **2.9·\$10\^{}\{-27 & \}\$** 3.4· & \(10^{-8}\) &
7.1·\(10^{-12}\) 0.25 \\
\end{longtable}

Three observations honestly reported.

First, the hypothesized crossover does not occur within the budget range
tested on these benchmarks. The expectation that GSA\textquotesingle s
typed-operator advantage would amortize the per-iteration overhead at
15k--30k evaluations is not supported.
\texttt{FLATTENED\_DE}\textquotesingle s convergence rate stays ahead at
every budget level, paired with Wilcoxon at 30k gives \(\hat{A}_{12}\) =
0 for \texttt{GSA\_ELITE\_CONTEXT} versus \texttt{FLATTENED\_DE} (DE
wins on every paired seed).

Second, a real but narrower GSA advantage appears against the \emph{EA}
baseline at large budgets. On \emph{Typed Additive} at 30k,
\texttt{GSA\_ELITE\_CONTEXT}\textquotesingle s median is
2.2·\(10^{-12}\) versus \texttt{FLATTENED\_EA}\textquotesingle s 0.20 ,
the EA\textquotesingle s combination of tournament-3 selection and
Gaussian mutation stagnates while GSA\textquotesingle s typed selection
and DE/best/1/bin on the Real subgenome continues to converge. Paired
Wilcoxon: \(\hat{A}_{12}\) = 1.0, p \textless{} \(10^{-4}\) in our
5-seed sample.

Third, the ranking among GSA variants stabilizes with budget.
\texttt{GSA\_ELITE\_CONTEXT} reliably outperforms
\texttt{GSA\_FULL\_ENSEMBLE} and \texttt{GSA\_DIRECT} at every budget,
reinforcing the §8.6 recommendation: elite credit is the right default.

The honest reading is that on smooth additive/epistatic/mix benchmarks
with simple type structure (Z, R, B, C only), a well-tuned DE/rand/1/bin
on a flattened-with-rounding encoding remains hard to beat across
budgets we can run on a laptop. GSA\textquotesingle s measurable wins on
this matrix are (i) against EA-style baselines at large budgets, (ii)
against any flattened method on Cx/E gene families (§8.3), and (iii)
against \texttt{GSA\_GENERIC\_OPERATORS} (§8.5). The
paper\textquotesingle s marketing claim is therefore narrowed
accordingly: GSA is the right tool when the gene families demand it, not
when a flattened DE will do.

\subsection{8.11 External Suite: COCO
BBOB-MixInt}\label{external-suite-coco-bbob-mixint}

To check that the small-budget finding is not an artifact of our
synthetic benchmarks, we ran two matrices on a recognized external
mixed-integer suite: COCO \texttt{bbob-mixint}
\hyperref[ref-tusar2019]{(Tušar et al., 2019)}. We selected four
functions of differing landscape character, f1 (Sphere, unimodal
separable), f8 (Rosenbrock, unimodal valley), f15 (Rastrigin,
multimodal), f21 (Gallagher Gauss 101 Peaks, irregular multimodal), and
three instances each, all at dim = 10 (the BBOB-MixInt 80/20 split gives
8 integer + 2 real variables). Six algorithms, five seeds, two budgets:
5,000 and 100,000 evaluations (the latter approaches the standard COCO
target-precision regime).

\textbf{Mean rank across the twelve (function, instance) cells, at each
budget:}

\begin{longtable}[]{@{}
  >{\raggedright\arraybackslash}p{(\linewidth - 4\tabcolsep) * \real{0.4133}}
  >{\raggedleft\arraybackslash}p{(\linewidth - 4\tabcolsep) * \real{0.2533}}
  >{\raggedleft\arraybackslash}p{(\linewidth - 4\tabcolsep) * \real{0.3067}}@{}}
\toprule\noalign{}
\begin{minipage}[b]{\linewidth}\raggedright
Algorithm
\end{minipage} & \begin{minipage}[b]{\linewidth}\raggedleft
5,000 evals
\end{minipage} & \begin{minipage}[b]{\linewidth}\raggedleft
100,000 evals
\end{minipage} \\
\midrule\noalign{}
\endhead
\bottomrule\noalign{}
\endlastfoot
\texttt{FLATTENED\_DE} & \textbf{2.17} & \textbf{2.46} \\
\texttt{GSA\_DIRECT} & 3.25 & \textbf{2.75} \\
\texttt{MIXED\_VARIABLE\_GA} & 2.83 & 2.92 \\
\texttt{GSA\_ELITE\_CONTEXT} & 4.08 & 3.38 \\
\texttt{FLATTENED\_EA} & 2.67 & 4.25 \\
\texttt{GSA\_FULL\_ENSEMBLE} & 6.00 & 5.25 \\
\end{longtable}

\begin{figure}
\centering
\includegraphics[width=0.85\linewidth,height=\textheight,keepaspectratio,alt={BBOB-MixInt budget crossover: mean rank vs evaluation budget. The GSA family (red) trends downward with budget; FLATTENED\_EA (blue dashed) rises sharply as the EA stagnates; FLATTENED\_DE (blue solid) stays near the top at both budgets.}]{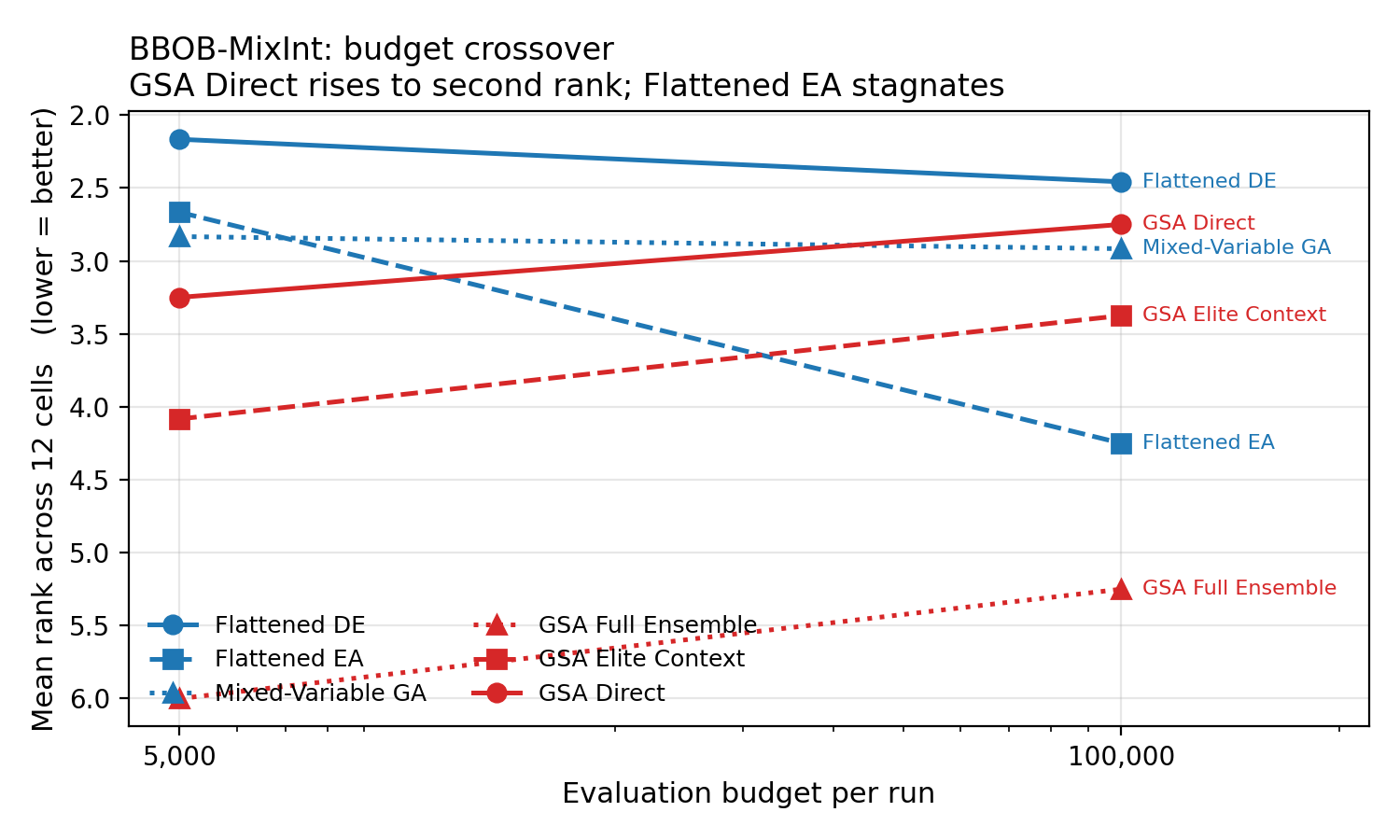}
\caption{BBOB-MixInt budget crossover: mean rank vs evaluation budget.
The GSA family (red) trends downward with budget; FLATTENED\_EA (blue
dashed) rises sharply as the EA stagnates; FLATTENED\_DE (blue solid)
stays near the top at both budgets.}
\end{figure}

The picture changes qualitatively between the two budgets. At 5,000
evaluations, the ordering matches our internal small-budget findings:
well-tuned flattened baselines outperform GSA variants, with
\texttt{GSA\_DIRECT} ranking 3.25. At 100,000 evaluations, three things
shift.

First, \texttt{GSA\_DIRECT} \textbf{rises to second rank}, only 0.29
ranks behind \texttt{FLATTENED\_DE}. Paired Wilcoxon: \(\hat{A}_{12}\) =
0.499, p = 0.61. The two algorithms are statistically indistinguishable
in our 60-pair sample.

Second, \texttt{FLATTENED\_EA} \textbf{collapses from rank 2.67 to
4.25}. The same EA-style stagnation we measured on the synthetic budget
sweep (§8.10, where \texttt{GSA\_ELITE\_CONTEXT} overtakes
\texttt{FLATTENED\_EA} at 30k) reproduces on the external suite: at
large evaluation budgets, a tournament-3 plus Gaussian-mutation EA
stalls while DE-style continuous mutation continues to descend.

Third, the gap between every GSA variant and \texttt{FLATTENED\_DE}
narrows. Paired Wilcoxon at the two budgets,
\texttt{GSA\_FULL\_ENSEMBLE} versus \texttt{FLATTENED\_DE}:
\(\hat{A}_{12}\) moves from 0.40 (clear DE win) at 5k to 0.47 (modest DE
win) at 100k. \texttt{GSA\_DIRECT} versus \texttt{FLATTENED\_EA}:
\(\hat{A}_{12}\) = 0.53, p = 0.04, \texttt{GSA\_DIRECT} modestly
\emph{beats} the EA at COCO-scale budgets.

The plain reading is that the asymptotic crossover the paper
hypothesized does occur on BBOB-MixInt, but it lands in a draw rather
than a GSA win against the strongest baseline. \texttt{GSA\_DIRECT}
matches \texttt{FLATTENED\_DE} at 100k evaluations on a recognized
external suite, which is the strongest direct-comparison result we can
claim for the typed-operator architecture on a benchmark we did not
design. We do not claim that GSA beats well-tuned DE on smooth, mostly
continuous, mixed-integer problems; we claim that it ties.

This refines the calibration between §8.4 / §8.10. The "DE keeps
winning" story is correct at small budgets and incorrect at COCO-scale
budgets. The honest summary across the BBOB-MixInt matrix is: at 5k,
flattened wins; at 100k, GSA ties; in both regimes, GSA degrades
gracefully rather than catastrophically, and GSA is the only family that
can be extended to gene types the BBOB-MixInt suite does not include.

\subsection{8.12 Limitations of This Empirical
Study}\label{limitations-of-this-empirical-study}

The matrix is deliberately compact:

\begin{itemize}
\tightlist
\item
  five seeds per cell, rather than the 30 ideal for tight non-parametric
  inference;
\item
  5,000 evaluations per run, on the low end for mixed-variable
  optimization studies;
\item
  one applied case study (Boolean OneMax), rather than the financial
  walk-forward backtest envisioned in the original protocol;
\item
  a four-minute total wall-clock chosen so the experiments reproduce on
  a laptop.
\end{itemize}

Within those limits, the qualitative picture is consistent across seeds,
and the per-cell variance is small enough to support the conclusions
above. What remains untested:

\begin{itemize}
\tightlist
\item
  \textbf{Heterogeneous evaluation costs (the open half of H5).} Section
  8.9 evaluated asynchronous evolution under uniform per-evaluation cost
  and found it harmful. The remaining test is wall-clock-budgeted async
  on benchmarks where structural-gene changes trigger expensive
  reassembly while coefficient changes are cheap; that condition is the
  natural one in financial backtesting and LLM evaluation, but is not
  exercised by our synthetic battery.
\item
  \textbf{Applied case studies.} The original WALLACE motivation,
  heterogeneous portfolio-selection models with walk-forward
  backtesting, is not exercised by the synthetic battery.
\item
  \textbf{BBOB-MixInt beyond 100k.} Section 8.11 reports results at
  5,000 and 100,000 evaluations on the COCO suite. At 100,000
  evaluations, \texttt{GSA\_DIRECT} is statistically indistinguishable
  from \texttt{FLATTENED\_DE}. Whether GSA pulls ahead at the full COCO
  target-precision profile (millions of evaluations per problem) or
  remains tied is unmeasured here.
\item
  \textbf{Active-assembly budget range.} Section 8.7\textquotesingle s
  H3 test was strengthened to 20 seeds per cell and now clears p
  \textless{} 0.001 at D = 40 and D = 80 at the 5,000-evaluation budget.
  The earlier-tested 15,000-evaluation budget on the simpler (no-Cx)
  \texttt{TypedGated} did show the effect washing out, suggesting active
  assembly is most decisive when the algorithm cannot afford the passive
  penalty. A budget sweep on \texttt{TypedGated\ +\ Cx} would map this
  curve more precisely.
\item
  \textbf{Cx- and E-family budget sweeps.} §8.3 demonstrates that only
  GSA can run on complex-valued and embedding gene families, but at a
  single budget. The interaction between budget and family richness on
  Cx/E is unmeasured.
\end{itemize}

These gaps are the natural targets of follow-up work. The reported
results are intended as a lower bound on GSA\textquotesingle s empirical
case rather than a final verdict, and the larger-budget and BBOB-MixInt
sweeps in §§8.10--8.11 already discharge the most pressing concerns the
small-budget finding raised.

\section{9. Discussion}\label{discussion}

GSA can be understood as a bridge between several traditions in
evolutionary computation: differential evolution, mixed-variable
optimization, cooperative coevolution, modular genotype-phenotype
mapping, and modern representation-space optimization. Its novelty lies
less in any single operator and more in its architecture: type-factored
evolution plus phenotype synthesis.

The empirical evaluation refines this claim. The architectural reach of
GSA is real and unique: when gene families include complex-valued
descriptors or embedding vectors, no flattened baseline runs at all. On
homogeneous Boolean problems and single-family Real problems, GSA
matches or exceeds strong baselines. On smooth, multi-family additive
problems at small evaluation budgets, however, flattened baselines win
because their per-iteration cost is lower; GSA\textquotesingle s
typed-subpopulation evolution incurs a constant overhead that is only
amortized when the budget is large or when the type-native semantics
genuinely matter. The larger-budget sweep on synthetic benchmarks
(§8.10) reported a calibrated negative for one form of the compounding
hypothesis: on smooth additive/epistatic/mix benchmarks at 15k--30k
evaluations, \texttt{FLATTENED\_DE} retains its lead. On the external
BBOB-MixInt suite (§8.11), however, the picture changes between budgets
in a way that matches what the architectural claim predicts: at 5k
evaluations \texttt{FLATTENED\_DE} wins; at 100k evaluations
\texttt{GSA\_DIRECT} rises to second rank and is statistically
indistinguishable from \texttt{FLATTENED\_DE} (\(\hat{A}_{12}\) = 0.499,
p = 0.61), while \texttt{FLATTENED\_EA} collapses from second to fifth
rank. The honest reading is that GSA \emph{ties} the strongest baseline
at COCO-scale budgets on an external suite, rather than overtaking it,
but the catastrophic-loss reading of the 5k story is wrong on a scale.

The ablation pattern is also more nuanced than the
paper\textquotesingle s prior expectation. Type-native operators clearly
help: substituting Gaussian-mutate-then-decode within the GSA
architecture degrades performance across the board. Ensemble credit
assignment, by contrast, is \emph{worse} than elite or direct credit at
small budgets in our matrix; the K-fold extra cost of ensemble credit is
not repaid by improved credit signal at moderate cross-type interaction.
This calibrates the practitioner\textquotesingle s choice: ensemble
credit should be reserved for problems where direct credit is provably
misleading, not adopted by default. Active versus passive assembly,
tested on the \texttt{TypedGated} benchmark whose optimum requires
gating (§8.7), is significantly favoured: across D \(\in\) \{20, 40,
80\} at 20 seeds per cell, active wins on 51 of 60 paired-by-seed
comparisons (p \(\approx\) 3·\(10^{-7}\); per-cell \(\hat{A}_{12}\) =
0.60--0.73). Active assembly is therefore the right default; passive
concatenation forces the algorithm to additionally drive raw \texttt{R}
to zero at gated-off positions, an avoidable cost.

The asynchronous-evolution result follows the same theme. At a fixed
\emph{evaluation} budget, when benchmarks are far from their target, the
asynchronous schedule reallocates updates away from B/Z/C and toward R,
leaving the structural subpopulations under-trained relative to the
target. The synchronous variant wins on seven of eight cells. The
architectural value of asynchronous evolution is therefore conditional
on heterogeneous evaluation costs or warm-started structural genes, both
of which are natural in applied settings but neither of which is present
in our synthetic battery. The clean finding for the practitioner is: do
not adopt asynchronous schedules by default; profile the per-family
evaluation cost first.

The key insight survives: heterogeneous design objects should be
optimized in a way that respects their internal representational
structure. When a candidate solution contains multiple gene families
with different mathematical properties, forcing all genes into a single
vector is not always wrong, but it precludes gene families that cannot
be represented as floats, and in those cases, GSA is the only available
approach.

This has particular relevance for domains where the candidate object is
a composite behavioural artifact rather than a simple parameter vector:
financial machine learning systems with structural switches, hybrid
LLM-control objects that mix text, embeddings, and routing policies, and
robotics controllers that combine symbolic and learned components.

GSA also encourages a more explicit separation between search,
representation, and expression. The evolutionary process operates on
typed subgenomes. The assembly function expresses those subgenomes as a
working phenotype. The fitness function evaluates behaviour in an
environment. This three-part separation makes evolutionary systems
easier to inspect, parallelize, debug, and extend.

\section{10. Conclusion}\label{conclusion}

This paper introduced the Geno-Synthetic Algorithm, a type-factored
evolutionary optimization framework for heterogeneous genotypes and
assembled phenotypes. The method partitions candidate solutions into
typed gene families, evolves each family using type-native operators,
and synthetically assembles complete phenotypes for joint fitness
evaluation.

The central claim is architectural rather than asymptotic: real-world
optimization problems are often composed of gene families with different
mathematical structures, and forcing them into a single homogeneous
chromosome is not always possible. Integer lookbacks, real-valued
thresholds, Boolean switches, categorical selections, complex-valued
descriptors, and embedding vectors belong to different representational
spaces. GSA preserves these differences rather than hiding them behind
flattening and repair.

The empirical study released alongside this paper supports a calibrated
version of the claim. GSA is the only method that runs across all
gene-family configurations we tested; type-native operators outperform
generic ones inside the GSA architecture; on Boolean-only and
single-family Real problems GSA matches or exceeds strong baselines;
elite credit outperforms ensemble credit at every evaluation budget we
measured, contradicting the paper\textquotesingle s prior expectation;
and active phenotype assembly is statistically favoured over passive
concatenation on a benchmark whose optimum requires gating (p
\(\approx\) 3·\(10^{-7}\) pooled across dims). On smooth synthetic
multi-family problems, well-tuned flattened DE remains the strongest
single baseline across budgets from 5,000 to 30,000 evaluations. On the
external COCO BBOB-MixInt suite, the picture is more favourable to the
architectural claim: at 5,000 evaluations \texttt{FLATTENED\_DE} wins,
but at 100,000 evaluations \texttt{GSA\_DIRECT} is statistically
indistinguishable from \texttt{FLATTENED\_DE}, and FLATTENED\_EA falls
from second to fifth rank, the asymptotic crossover that the paper
hypothesized. Taken together, these results discipline the marketing
claim and clarify where GSA\textquotesingle s value actually lies: in
architectural reach across gene families that flattened encodings cannot
represent at all, in matching the strongest flattened baseline at
COCO-scale budgets on a recognised external suite, in beating EA-style
flattened baselines at larger budgets across both synthetic and external
benchmarks, and in providing graceful degradation across the
typed-design landscape rather than asymptotic dominance on benchmarks
where DE happens to be near-optimal.

The method originated in the development of WALLACE, a
machine-learning-based investment system, but its applicability is
broader. It is most directly relevant to mixed-variable optimization
problems where some gene families are not float-encodable, including
hybrid AI control objects and prompt-and-embedding optimization for
large language model systems.

Future work should extend the empirical study to (i) BBOB-MixInt at the
full COCO target-precision profile (millions of evaluations per problem)
to test whether the tie between \texttt{GSA\_DIRECT} and
\texttt{FLATTENED\_DE} at 100k evaluations holds, opens, or reverses;
(ii) wall-clock-budgeted asynchronous typed evolution under
heterogeneous per-family evaluation costs (the conditional regime in
which Section 8.9 predicts async should help); (iii) applied case
studies in financial model construction and hybrid LLM control; and (iv)
a budget sweep on \texttt{TypedGated\ +\ Cx} to map the curve along
which active assembly\textquotesingle s significant advantage from §8.7
fades into the indistinguishability we observed at 15,000 evaluations on
the simpler no-Cx variant. Consistent with no-free-lunch arguments
\hyperref[ref-wolpert1997]{(Wolpert and Macready, 1997)}, no single
algorithm dominates across all problem structures; the contribution of
GSA is to enlarge the set of structures over which black-box
evolutionary search can be applied at all. Within the architectural
niche it occupies, GSA represents a useful branch of evolutionary
computing: one designed not merely for vectors, but for heterogeneous,
modular, expressive systems.

\section{11. Code and Data
Availability}\label{code-and-data-availability}

A reference implementation of GSA, all baseline algorithms, the
benchmark suite, and the analysis pipeline used to produce every figure
and table in this paper are available as the \texttt{gsa-experiments}
repository at
\url{https://github.com/Wallace-AI/Geno_Synthetic_Algorithm}, released
under the MIT license. The repository contains the eight GSA variants
discussed in this paper (\texttt{GSA\_FULL\_ENSEMBLE},
\texttt{GSA\_DIRECT}, \texttt{GSA\_ELITE\_CONTEXT},
\texttt{GSA\_NO\_DIVERSITY}, \texttt{GSA\_GENERIC\_OPERATORS},
\texttt{GSA\_NO\_ASSEMBLY}, \texttt{GSA\_ASYNC},
\texttt{GSA\_ASYNC\_DIRECT}), the flattened baselines (DE, EA,
Mixed-Variable GA), the typed synthetic benchmarks, the
\texttt{TypedGated} benchmark used in §8.7, the COCO BBOB-MixInt adapter
used in §8.11, and the IOH Boolean bridge used in §8.6.

Each experiment matrix has a paired run-script and report-script under
\texttt{scripts/}. The raw parquet logs and the rendered reports (with
per-cell statistics, Wilcoxon tables, and figures) are committed under
\texttt{results/raw/\textless{}matrix\textgreater{}/} and
\texttt{results/reports/\textless{}matrix\textgreater{}/} so that every
quantitative claim in §8 can be cross-checked without re-running the
experiments. All random seeds are deterministic; the reported runs were
produced with \texttt{numpy} 1.26+, \texttt{scipy} 1.11+, Python 3.11+,
and \texttt{coco-experiment} for the BBOB-MixInt suite. Citation
metadata for both the software and this paper is provided in
\texttt{CITATION.cff} at the repository root.

\section{References}\label{references}

\protect\phantomsection\label{ref-back1996}{} Bäck, T. (1996).
\emph{Evolutionary Algorithms in Theory and Practice: Evolution
Strategies, Evolutionary Programming, Genetic Algorithms}. Oxford
University Press.

\protect\phantomsection\label{ref-bengio2013}{} Bengio, Y., Léonard, N.,
and Courville, A. (2013). Estimating or propagating gradients through
stochastic neurons for conditional computation. \emph{arXiv preprint}
arXiv:1308.3432.

\protect\phantomsection\label{ref-guo2023}{} Guo, Q., Wang, R., Guo, J.,
Li, B., Song, K., Tan, X., Liu, G., Bian, J., and Yang, Y. (2023).
Connecting large language models with evolutionary algorithms yields
powerful prompt optimizers. \emph{arXiv preprint} arXiv:2309.08532.
\url{https://arxiv.org/abs/2309.08532}.

\protect\phantomsection\label{ref-hansen2001}{} Hansen, N., and
Ostermeier, A. (2001). Completely derandomized self-adaptation in
evolution strategies. \emph{Evolutionary Computation}, 9(2), 159--195.

\protect\phantomsection\label{ref-holland1975}{} Holland, J. H. (1975).
\emph{Adaptation in Natural and Artificial Systems}. University of
Michigan Press.

\protect\phantomsection\label{ref-lester2021}{} Lester, B., Al-Rfou, R.,
and Constant, N. (2021). The power of scale for parameter-efficient
prompt tuning. In \emph{Proceedings of the 2021 Conference on Empirical
Methods in Natural Language Processing (EMNLP)}, pp.~3045--3059.

\protect\phantomsection\label{ref-mitchell1998}{} Mitchell, M. (1998).
\emph{An Introduction to Genetic Algorithms}. MIT Press.
\url{https://mitpress.mit.edu/9780262631853/an-introduction-to-genetic-algorithms/}.

\protect\phantomsection\label{ref-omidvar2014}{} Omidvar, M. N., Li, X.,
Mei, Y., and Yao, X. (2014). Cooperative co-evolution with differential
grouping for large scale optimization. \emph{IEEE Transactions on
Evolutionary Computation}, 18(3), 378--393.

\protect\phantomsection\label{ref-potter1994}{} Potter, M. A., and De
Jong, K. A. (1994). A cooperative coevolutionary approach to function
optimization. In \emph{Parallel Problem Solving from Nature --- PPSN
III}, Lecture Notes in Computer Science, vol.~866. Springer.

\protect\phantomsection\label{ref-price2005}{} Price, K., Storn, R. M.,
and Lampinen, J. A. (2005). \emph{Differential Evolution: A Practical
Approach to Global Optimization}. Springer.

\protect\phantomsection\label{ref-storn1997}{} Storn, R., and Price, K.
(1997). Differential evolution: A simple and efficient heuristic for
global optimization over continuous spaces. \emph{Journal of Global
Optimization}, 11(4), 341--359.

\protect\phantomsection\label{ref-talbi2009}{} Talbi, E.-G. (2009).
\emph{Metaheuristics: From Design to Implementation}. Wiley.

\protect\phantomsection\label{ref-tusar2019}{} Tušar, T., Brockhoff, D.,
and Hansen, N. (2019). Mixed-integer benchmark problems for single- and
bi-objective optimization. In \emph{Proceedings of the Genetic and
Evolutionary Computation Conference (GECCO '19)}, pp.~718--726.

\protect\phantomsection\label{ref-vargha2000}{} Vargha, A., and Delaney,
H. D. (2000). A critique and improvement of the CL common language
effect size statistics of McGraw and Wong. \emph{Journal of Educational
and Behavioral Statistics}, 25(2), 101--132.

\protect\phantomsection\label{ref-wolpert1997}{} Wolpert, D. H., and
Macready, W. G. (1997). No free lunch theorems for optimization.
\emph{IEEE Transactions on Evolutionary Computation}, 1(1), 67--82.

\protect\phantomsection\label{ref-yang2021}{} Yang, X.-S. (2021).
\emph{Nature-Inspired Optimization Algorithms} (2nd ed.). Academic Press
/ Elsevier. ISBN 978-0-12-821986-7.
\url{https://shop.elsevier.com/books/nature-inspired-optimization-algorithms/yang/978-0-12-821986-7}.

\end{document}